%% file: MAIN.tex
\documentclass[]{interact}
\usepackage{epstopdf}
\usepackage[caption=false]{subfig}
\usepackage[overload]{empheq}

\usepackage[numbers,sort&compress]{natbib}
\bibpunct[, ]{[}{]}{,}{n}{,}{,}
\renewcommand\bibfont{\fontsize{10}{12}\selectfont}
\makeatletter
\def\NAT@def@citea{\def\@citea{\NAT@separator}}
\makeatother

\theoremstyle{plain}

\theoremstyle{definition}

\theoremstyle{remark}

\usepackage{setspace}
\usepackage[latin1]{inputenc}
\usepackage{amsmath}
\usepackage{amsfonts}
\usepackage{graphicx}
\usepackage{subfig}
\usepackage{float}
\usepackage{siunitx}
\usepackage{listings}
\usepackage{pdfpages}
\usepackage{keyval}
\usepackage{eso-pic}
\usepackage{atbegshi}
\usepackage{pdflscape}
\usepackage{booktabs}
\usepackage{multirow}
\usepackage{rotating}
\usepackage{url}
\usepackage{array}
\usepackage{fancyhdr}
\usepackage{enumitem}
\usepackage{lipsum}
\usepackage{adjustbox}

\usepackage[np]{numprint}
\npdecimalsign{\ensuremath{.}}

\usepackage{caption}

\usepackage{algorithm,algpseudocode}
\usepackage{tikz,tikz-3dplot}
\usetikzlibrary{arrows.meta,angles,decorations.pathmorphing,patterns}

\usepackage{bm}

\usepackage{etoolbox} 

\DeclareMathOperator{\diag}{diag}

\newcommand\Var[1]{\sigma_{#1}^2}
\newcommand\Cov[2]{\sigma_{#1\,#2}}

\input{mymacros}

\begin{document}


\title{Disturbance-aware minimum-time planning strategies for motorsport vehicles with probabilistic safety certificates}

\author{Martino Gulisano\textsuperscript{a}, Matteo Masoni\textsuperscript{a}, \name{Marco Gabiccini\textsuperscript{a,*} and Massimo Guiggiani\textsuperscript{a}
\thanks{* Contact: Marco Gabiccini. Email: marco.gabiccini@unipi.it}
}\affil{
\textsuperscript{a} Dipartimento di Ingegneria Civile e Industriale,	Universit\`{a} di Pisa, Pisa, Italy.}
}

\maketitle

\begin{abstract}
This paper presents a disturbance-aware framework that embeds robustness into minimum-lap-time trajectory optimization for motorsport. Two formulations are introduced. (i) Open-loop, horizon-based covariance propagation uses worst-case uncertainty growth over a finite window to tighten tire-friction and track-limit constraints. (ii) Closed-loop, covariance-aware planning incorporates a time-varying LQR feedback law in the optimizer, providing a feedback-consistent estimate of disturbance attenuation and enabling sharper yet reliable constraint tightening. Both methods yield reference trajectories for human or artificial drivers: in autonomous applications the modelled controller can replicate the on-board implementation, while for human driving accuracy increases with the extent to which the driver can be approximated by the assumed time-varying LQR policy.  Computational tests on a representative Barcelona-Catalunya sector show that both schemes meet the prescribed safety probability, yet the closed-loop variant incurs smaller lap-time penalties than the more conservative open-loop solution, while the nominal (non-robust) trajectory remains infeasible under the same uncertainties.
By accounting for uncertainty growth and feedback action during planning, the proposed framework delivers trajectories that are both performance-optimal and probabilistically safe, advancing minimum-time optimization toward real-world deployment in high-performance motorsport and autonomous racing.
\end{abstract}

\begin{keywords}
Minimum lap-time trajectory planning; stochastic vehicle dynamics; probabilistic safety certificates.
\end{keywords}

\input{intro}
\input{stochastic_vehicle_dynamics}
\input{open_loop_planning}
\input{closed_loop_planning}
\input{performance_comparison}

\input{conclusions}

\section*{Acknowledgement}

The authors would like to thank Dr. Lorenzo Bartali for his significant contributions to the initial version of the optimization framework developed and utilized in this paper.

\section*{Disclosure statement}

No potential conflict of interest was reported by the author(s).

\section*{Funding}

This work is supported by project PRIN 2022 PNRR ``Global Stability of road vehicle motion - STAVE'' CUP I53D23005670001.

%
%
%
%
%
%
%

\bibliographystyle{tfnlm}
\bibliography{Disturbance-Aware-Planning}

%
%
%


%
%
%
%

\end{document}

%% file: mymacros.tex
\newcommand{\De}{\Delta}

\newcommand{\Si}{\Sigma}

\newcommand{\Om}{\Omega}

\newcommand{\al}{\alpha}
\newcommand{\be}{\beta}

\newcommand{\ga}{\gamma}
\newcommand{\de}{\delta}

\newcommand{\sig}{\sigma}

\newcommand{\bsig}{\boldsymbol{\sig}}  
\newcommand{\bSi}{\boldsymbol{\Si}}
\newcommand{\bOm}{\boldsymbol{\Om}}  
\newcommand{\bxi}{\boldsymbol{\xi}}  
\newcommand{\bmu}{\boldsymbol{\mu}}
\newcommand{\bPhi}{\boldsymbol{\Phi}}
\newcommand{\dbPhi}{\dot{\bPhi}}

\newcommand{\bPsi}{\boldsymbol{\Psi}}
\newcommand{\bPsimu}{\bPsi^{\mu}}%
\newcommand{\bPsiP}{\bPsi^{\text{P}}}%
\newcommand{\bPsiPCL}{\bPsi^{\text{P,CL}}}%

\newcommand{\dbmu}{\dot{\bmu}}

\newcommand{\hbmu}{\hat{\bmu}}
\newcommand{\hbu}{\hat{\bu}}
\newcommand{\tbu}{\tilde{\bu}}

\newcommand{\bbmu}{\bar{\bmu}}
\newcommand{\bbu}{\bar{\bu}}
\newcommand{\bbsig}{\bar{\bsig}}

\newcommand{\hbz}{\hat{\bz}}

\newcommand{\hbA}{\hat{\bA}}
\newcommand{\hbB}{\hat{\bB}}
\newcommand{\tbB}{\tilde{\bB}}

\newcommand{\hbK}{\hat{\bK}}
\newcommand{\tbK}{\tilde{\bK}}

\newcommand{\tbP}{\tilde{\bP}}
\newcommand{\tbA}{\tilde{\bA}}

\newcommand{\bn}{\boldsymbol{n}}      

\newcommand{\bu}{\boldsymbol{u}}

\newcommand{\bw}{\boldsymbol{w}}
\newcommand{\bx}{\boldsymbol{x}}     

\newcommand{\bz}{\boldsymbol{z}}

\newcommand{\bA}{\boldsymbol{A}}     
\newcommand{\bB}{\boldsymbol{B}}

\newcommand{\bH}{\boldsymbol{H}}
\newcommand{\bI}{\boldsymbol{I}}

\newcommand{\bK}{\boldsymbol{K}}

\newcommand{\bP}{\boldsymbol{P}}
\newcommand{\bQ}{\boldsymbol{Q}}
\newcommand{\bR}{\boldsymbol{R}}
\newcommand{\bS}{\boldsymbol{S}}

\newcommand{\bW}{\boldsymbol{W}}

\newcommand{\bzero}{\boldsymbol{0}}

\newcommand{\dbx}{\dot{\bx}}     

\newcommand{\dbP}{\dot{\bP}}

\newcommand{\barbx}{\bar{\bx}}

\newcommand{\bbR}{\mathbb{R}}    

\newcommand{\calN}{\mathcal{N}}

\newcommand{\calI}{\mathcal{I}}

\DeclareMathOperator{\Prob}{Pr}

\newcommand{\dd}{\text{{\sl d}}}        

\newcommand{\na}{\nabla}

\def\pd{\partial}

\newcommand{\Ipos}[1]{\mathbb{I}_{\geq 0}(#1)}
\newcommand{\Ineg}[1]{\mathbb{I}_{< 0}(#1)}

\newcommand{\SIpos}[1]{\mathbb{SI}_{\geq 0}(#1)}
\newcommand{\SIneg}[1]{\mathbb{SI}_{ < 0}(#1)}



%% file: intro.tex
\section{Introduction}
\label{sec:intro}

Minimum lap time optimization is a fundamental tool in the motorsport field, enabling the synthesis of optimal trajectories and control profiles that push performance to its limits. These optimal references are widely used both offline, for vehicle setup and strategy development, and online, as feedforward inputs to advanced driver-assistance and autonomous systems.
However, despite their optimality under nominal assumptions, trajectories produced by state-of-the-art planners often lie critically close to physical and safety constraints---such as tire friction limits and collision boundaries---rendering them extremely sensitive to disturbances and modeling inaccuracies. Consequently, these ideal references may prove difficult or unsafe to follow, even for expert drivers, limiting their practical usability.

This fragility highlights a critical gap: current minimum lap time formulations rarely embed robustness explicitly.
Consequently, resulting trajectories lack reliability under uncertainties arising from mismatches between modeling assumptions and real operating conditions.
Addressing this shortcoming is essential to bridge the gap between theoretical optimality and real-world feasibility in high-performance motorsport applications.

\subsection{Related work}
In recent years, comprehensive analyses have been conducted on minimum-lap-time optimization for motorsport vehicles, covering fixed- and free-trajectory formulations~\cite{Veneri:FreetrajectoryQuasisteadystateOptimalcontrol:2020, Lovato:ThreedimensionalFixedtrajectoryApproaches:2022, Lovato:ThreedimensionalFreetrajectoryQuasisteadystate:2022}, comparing direct and indirect solution techniques~\cite{DalBianco:ComparisonDirectIndirect:2019, Bertolazzi:DirectIndirectApproach:2025}, and contrasting serial and parallel solver frameworks~\cite{Biniewicz:QuasisteadystateMinimumLap:2024, Bartali:SchwarzDecompositionParallel:2024, Bartali:ConsensusbasedAlternatingDirection:2024}.

Despite these advances, most planning frameworks still fail to incorporate robustness in the planning phase itself. Piccinini et al.~\cite{Piccinini:HowOptimalMinimumtime:2024} provide a direct comparison between an offline minimum-lap-time optimal control problem (MLT-OCP) and an online Artificial Racing Driver (ARD) that controls the very same high-fidelity vehicle model. They show that, by leveraging a physics-driven structure and a novel g-g-v performance constraint, ARD can achieve lap times within a few milliseconds of the offline benchmark and generalize to unseen circuits even under unmodeled mass variations. However, their work remains focused on quantifying the execution gap - how ARD mitigates local tracking errors - rather than on embedding disturbance handling directly into the trajectory planner.

The omission of explicit disturbance modeling at the planning stage can critically undermine constraint satisfaction: time-optimal planners typically produce trajectories that closely approach safety boundaries (e.g., collision avoidance or tire-grip limits), where even minor perturbations may lead to violations and thus severely compromise system safety.
While the use of fixed, heuristically defined safety margins around constraint sets may provide a nominal safeguard, such an approach lacks formal guarantees under uncertainty and typically leads to overly conservative and suboptimal solutions. This motivates the need for planning methods that explicitly account for uncertainty and provide quantifiable safety assurances.

In diverse fields, a variety of strategies has been proposed to tackle planning under uncertainty.
In orbital mechanics, uncertainty in state estimation necessitates a probabilistic framework for modeling interactions and potential close approaches among natural and artificial celestial bodies.
A foundational reference in this context is~\cite{Tapley:StatisticalOrbitDetermination:2004}, where statistical orbit determination techniques are employed to assess and mitigate collision risks. In robotic motion planning, uncertainty-aware techniques are crucial to ensure safe navigation in environments populated with obstacles. One of the earliest contributions proposing a probabilistic representation of uncertainty is the chance-constrained framework in~\cite{Blackmore:ChanceConstrainedOptimalPath:2011}, which plans over the predicted distribution of the system state to ensure that the probability of constraint violation remains below a specified threshold.

Within the Model Predictive Control (MPC) paradigm, recent works have incorporated probabilistic safety guarantees. Notably, the methods presented in~\cite{Gao:CollisionfreeMotionPlanning:2023},~\cite{Zhang:RobustifiedTimeoptimalPointtopoint:2025}, and~\cite{Zhang:RobustifiedTimeoptimalCollisionfree:2024} introduce stochastic MPC frameworks for autonomous mobile platforms, where process noise is explicitly modeled and closed-loop tracking performance is maintained via either pre-computed or optimized feedback gains. The approach in~\cite{Gao:CollisionfreeMotionPlanning:2023} additionally proposes a zero-order optimization scheme to include the feedback gain directly in the optimal control problem, mitigating the growth of uncertainty over the planning horizon.
In the domain of chemical engineering, robust MPC approaches have been proposed to address uncertainty in industrial settings. For example,~\cite{Krog:SimpleFastRobust:2024} presents a heuristic method based on $n$-step-ahead uncertainty predictions, which are used to compute constraint tightening margins for the control of a polymerization reactor.

In some contexts, uncertainty arises from partial or imprecise knowledge of system parameters. This has led to a significant body of work on robust trajectory planning, particularly for unmanned aerial vehicles (UAVs), with the aim of minimizing sensitivity to state and input variations. For instance,~\cite{Brault:RobustTrajectoryPlanning:2021} and~\cite{Giordano:TrajectoryGenerationMinimum:2018} propose tube-based and sensitivity-aware optimization frameworks that explicitly account for both input and closed-loop state sensitivity in the planning phase. Further developments integrate observability-aware planning into a unified multi-step optimization framework, as demonstrated in~\cite{Bohm:COPControlObservabilityaware:2022}.

Following an alternative approach, the largest Lyapunov exponent (LLE) has been proposed as an indicator of local stability~\cite{Meng:AnalysisGlobalCharacteristics:2022}. The LLE quantifies the average exponential rate at which infinitesimal perturbations off a fiducial trajectory grow or decay---thus providing a direct measure of how deviations propagate through the system.
Applications of this method are particularly relevant to study chaotic vessel motions~\cite{McCue:UseLyapunovExponents:2011} and rotorcraft dynamics~\cite{Tamer:StabilityNonlinearTimeDependent:2016,Cassoni:RotorcraftStabilityAnalysis:2024}, where the primary concern is understanding the long-term evolution of the system and determining its asymptotic behavior.
Although Lyapunov-based indicators have been applied in vehicular contexts as well~\cite{Sadri:StabilityAnalysisNonlinear:2013,Meng:AnalysisGlobalCharacteristics:2022}, their utility in planning is limited by the fact that asymptotic stability -- while theoretically informative -- often lacks operational relevance. A point along the planned trajectory may belong to the basin of attraction of a stable equilibrium, but if that equilibrium lies far ahead in time or outside the operational domain (e.g., off track), its practical relevance is questionable.

In our setting, the key question is whether perturbations occurring in unstable segments of the nominal trajectory can be handled via feedback or short-horizon replanning to keep the vehicle within safety-critical limits, such as tire forces and collision boundaries. This reframes the problem: rather than long-term convergence, we evaluate whether the planned trajectory ensures short-term stabilizability under uncertainty while maintaining constraint satisfaction.

\subsection{Paper's contributions and organization}
We tackle the challenge of embedding robustness against disturbances directly into minimum-lap-time planning and propose two complementary methods. In the first one---the open-loop \emph{horizon-based} covariance propagation---at each discretization point, we propagate the state covariance forward over a fixed horizon and tighten all path constraints against the maximum covariance growth. By back-offing constraints using this worst-case covariance, the resulting trajectory maximizes robustness without relying on feedback during planning.

In the second one -- the closed-loop covariance-aware planning, we integrate a time-varying LQR feedback policy into the planning process to realistically tame uncertainty growth. First, a nominal time-optimal trajectory is computed under deterministic dynamics. Second, an LQR controller is designed to stabilize that nominal path. Finally, we re-solve the planning problem - together with the feedback gains - so that, along this new robust trajectory, the closed-loop covariance propagation satisfies tightened constraints via a Lyapunov-based formulation. These two approaches offer a balance between computational simplicity and realistic, feedback-informed robustness, laying the groundwork for safe, high-performance lap-time planning under uncertainty in the motorsport context.

The rest of the paper is organized as follows.
In Sec.~\ref{sec:svdf} we introduce our stochastic dynamics framework, recalling the single-track vehicle model with nonlinear tires used for planning, and formulate both the continuous-time planning problem and its discrete-time counterpart via direct-collocation. Here, we also derive the probabilistic constraint back-off formulation.
Sec.~\ref{sec:open_loop_planning} then outlines the open-loop horizon-based covariance-propagation planning approach, while Sec.~\ref{sec:closed_loop_planning} presents the closed-loop, covariance-aware planner. In Sec.~\ref{sec:performance_comparison} we compare both methods on a representative track scenario, reporting parameter-sensitivity studies, analyzing the influence of different sets of constraints, and performance trade-offs.
Finally, Sec.~\ref{sec:conclusions} draws conclusions and offers directions for future work.
  

%% file: stochastic_vehicle_dynamics.tex
\section{Stochastic vehicle dynamic framework}
\label{sec:svdf}

\subsection{Vehicle model}
\label{sec:vehicle_model}
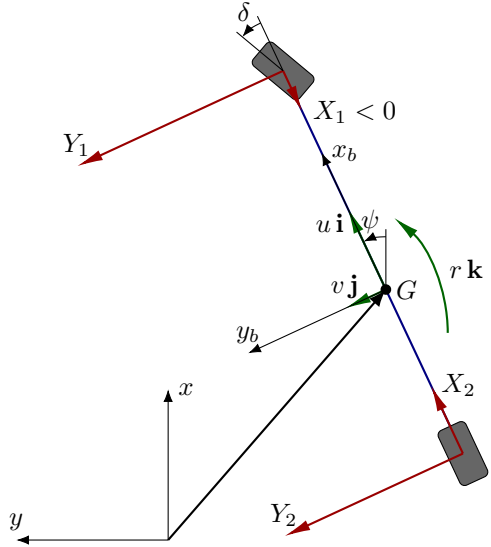
\begin{figure}
	\centering
	\input{TikZ_Sources/tikz_vehicle_model.tex}
	\caption{Single Track model.}
	\label{fig:vehicle_model}
\end{figure}

The vehicle model employed in this study is a Single Track model featuring nonlinear axle characteristics. A schematic representation of the Single Track model is shown in Fig.~\ref{fig:vehicle_model}. The lateral forces $Y_1$ and $Y_2$ are computed using a Pacejka's Magic Formula, which depends on the total vertical load acting on the axle and the axle's apparent slip angle.
The longitudinal forces $X_1$ and $X_2$ are provided as inputs to the system. Specifically, the first two components of the input vector $\bu$ correspond to acceleration and braking forces, while the third component represents the steering angle of the front wheels. This input formulation facilitates the distribution of braking forces between the front and rear axles, while assigning acceleration exclusively to the rear axle, consistent with the rear-wheel-drive configuration of the modeled vehicle.

By treating the longitudinal forces as inputs, the longitudinal dynamics can be explicitly solved, allowing the computation of vertical load transfers prior to their use in the expression of the lateral forces. Consequently, the time derivative of the state becomes an explicit function of the state vector $\bx$ and the input vector $\bu$, as follows:
\begin{equation}
	\dbx(t) = f(\bx(t), \bu(t))
\end{equation}

The state vector $\bx$ has six components and includes the three kinematic quantities $u$, $v$, and $r$, along with the position $\left(x_G, y_G\right)$ and orientation $\psi$ of the barycentrical reference frame. As illustrated in Fig.~\ref{fig:vehicle_model}, $u$ and $v$ represent the longitudinal and lateral velocity of the CoM w.r.t its reference frame, while $r$ represents vehicle's yaw rate.

\subsection{Stochastic vehicle dynamics}
\label{sec:stochastic_vehicle_dynamics}

We model the \emph{perturbed vehicle dynamics} as the following nonlinear continuous-time random dynamical system
\begin{align}
\dbx(t) = f(\bx(t), \bu(t)) + \bw(t),
\end{align}
where $\bu(t)$ are the deterministic control inputs, $\bw(t)$ is additive Gaussian white noise with zero mean and known covariance $\bQ(t)$, i.e. $\bw(t) \sim \calN(\bzero, \bQ(t))$, and $\bx(t)$ is the state vector.
Assuming a \emph{first-order} approximation for the disturbance propagation rule, $\bx(t)$ results in a Gaussian distribution with mean $\bmu(t)$ and covariance $\bP(t)$, so that $\bx(t)~\sim~\calN(\bmu(t), \bP(t))$.

Due to the symmetry of the probability density function (pdf) with respect to $\bmu(t)$, it is possible to represent the time evolution of the pdf as: i) the deterministic evolution of the mean $\bmu(t)$
\begin{align}\label{eq:meandynamics}
\dbmu(t) = f(\bmu(t), \bu(t)),
\end{align}
and ii) the time evolution of the state covariance matrix $\bP(t)$ along $\bmu(t)$, which can be expressed by the \emph{Lyapunov matrix differential equation}
\begin{align}\label{eq:dP}
\dbP(t) = \bA(t) \bP(t) + \bP(t) \bA^T(t) + \bQ(t),\quad \bP(0) = \bP_0 = \bP_0^T.
\end{align}
In~\eqref{eq:dP}, $\bP_0$ is the \emph{initial} state covariance and $\bA(t)$ is the usual shorthand notation for the Jacobian along the mean trajectory $\bmu(t)$, that is $\bA(t)=\frac{\pd f(\bmu, \bu)}{\pd \bmu}$.

As can be readily verified by differentiation~\cite{Gajic:LyapunovMatrixEquation:2010}, the analytical solution of~\eqref{eq:dP} has the form
\begin{align}\label{eq:P_STM}
\bP(t) = \bPhi(t,t_0) \bP_0 \bPhi^T(t,t_0)+\int_{t_0}^{t} \bPhi(t,\tau) \bQ(\tau) \bPhi^T(t,\tau) \dd \tau \quad \bP(0)=\bP_0,
\end{align}
where $\bPhi(t,t_0)$ is the \emph{state transition matrix}. This matrix encodes the evolution from $t_0$ to $t$ of a perturbation w.r.t. to the mean trajectory $\bmu(t)$. In symbols, $\barbx(t) = \bPhi(t,t_0) \barbx(t_0)$, with $\barbx(\cdot) =\bx(\cdot) - \bmu(\cdot) $. In turn, the evolution of $\bPhi(t,t_k)$ from a generic $t_k$ to $t$ is driven by the following differential equation
\begin{align}\label{eq:STM}
\dbPhi(t,t_k) = \bA(t)\bPhi(t,t_k), \quad \bPhi(t_k, t_k) = \bI.
\end{align}
In geometric and orbital mechanics, see e.g.~\cite{Maruskin:DynamicalSystemsGeometric:2018} or~\cite{Tapley:StatisticalOrbitDetermination:2004}, the usual choice for statistical trajectory determination is to employ~\eqref{eq:meandynamics} along with~\eqref{eq:P_STM} and~\eqref{eq:STM}. In our case, since we use collocation integrators for stochastic trajectory planning, we follow a more direct approach -- motivated by~\cite{Gillis:PracticalMethodsApproximate:2015} -- by directly employing~\eqref{eq:meandynamics} and~\eqref{eq:dP}.

Accordingly, the continuous-time stochastic trajectory planning can be framed as the following nonlinear optimal control problem
\begin{subequations}\label{eq:OCP}
\begin{align}
	\underset{\bmu(t), \bu(t), \bP(t)}{\text{minimize}} \quad & J(\bmu(t), \bu(t), \bP(t)) \label{eq:OCPcost} \\
	\text{s.t.} \quad \dbmu(t)           &= f(\bmu(t), \bu(t)) \label{eq:OCPdyn} \\
	\phantom{\text{s.t.} \quad} \bmu(0)  &= \bmu_0 \label{eq:OCPdynIC} \\
	\phantom{\text{s.t.} \quad} \dbP(t) &= \bA(t) \bP(t) + \bP(t) \bA^T(t) + \bQ(t) \label{eq:OCPdP} \\
	\phantom{\text{s.t.} \quad} \bP(0)  &= \bP_0 \succeq 0 \label{eq:OCPdPIC} \\ 
	\phantom{\text{s.t.} \qquad} 0&       \geq h_i(\bmu(t), \bu(t))
	+ \be_i(\bmu(t), \bu(t), \bP(t)),
	\quad i \in \calI \label{eq:OCPconstraints}
\end{align}
\end{subequations}
The cost function $J$ in~\eqref{eq:OCPcost} depends on the mean $\bmu(t)$, the controls $\bu(t)$ and the state covariance $\bP(t)$. The mean dynamics is expressed by~\eqref{eq:OCPdyn} with~\eqref{eq:OCPdynIC}, and the covariance dynamics is expressed by~\eqref{eq:OCPdP} with~\eqref{eq:OCPdPIC}. In eq.~\eqref{eq:OCPconstraints} the \emph{back-off terms} $\be_i$ account for the disturbances and serve the purpose of obtaining deterministic safety margins directly on $\bmu(t)$. $\calI$ is the set of indices defining the inequality constraints. The back-off terms stem from a linearization around the mean of the original \emph{chance constraint} on $\bx(t)$ expressed by $\Prob \{h_i(\bx) \leq 0\}\geq p$, where $p$ is the confidence level of constraint satisfaction. Explicitly, the back-off terms are given by
$\be_i = \gamma \sig_i$. The coefficient $\gamma = \Phi^{-1}(p)$ is the quantile function, where $\Phi(z)=\Prob \{Z\leq z\}$ is the cumulative distribution function (cdf) of a standard normal distribution $Z \sim \calN(0,1)$, and acts as a \emph{tuning knob}: the greater the confidence level $p$ required, the higher the gain $\ga$\footnote{For example, with $p=0.84$ $\ga = 1.0$, with $p=0.97$ $\ga = 2.0$, with $p=0.99$ $\ga = 3.0$}. The term $\sig_i = \big[\na^T_{\bx} h_i(\bmu) \bP(t) \na_{\bx} h_i(\bmu)\big]^{\frac{1}{2}}$ represents the standard deviation of the constraint $h_i(\bx)$ linearized around the mean, i.e. of random variable $h_i(\bmu)+\na^T_{\bx} h_i(\bmu) (\bx - \bmu)$, and follows immediately from the propagation rule of covariance.

\subsection{Discretization via direct collocation}
\label{sec:discretization}
The nonlinear optimal control problem~\eqref{eq:OCP} can be discretized by applying a suitable collocation integrator obtaining the following nonlinear program (NLP)
\begin{subequations}\label{eq:DOCP}
\begin{alignat}{3}
\underset{\bmu_k,\bxi_k, \bu_k, \bP_k,\bz_k}{\text{minimize}} \,
& & & J_k(\bmu_k,\bxi_k, \bu_k) & & \label{eq:DOCPcost} \\
\hspace*{-2.0 cm}\text{s.t.} \quad
& \bzero      & = & \; \bPsimu_k(\bmu_{k-1},\bmu_k,\bxi_k, \bu_k,\bz_k),
& \quad & k = 1,\ldots, N \label{eq:DOCPdyn} \\
& \bmu_0      & = & \; \bar{\bmu}_0
& & \label{eq:DOCPdynIC} \\
& \bzero      & = & \; \bPsiP_k(\bmu_k,\bxi_k, \bu_k, \bP_{k-1},\bP_k,\bSi_k,\bz_k),
& \quad & k = 1,\ldots, N \label{eq:DOCPdP} \\
& \bP_0       & = & \; \bar{\bP}_0 \succeq 0
& & \label{eq:DOCPdPIC} \\
& \bzero      & = & \; \bOm_k(\bmu_k,\bxi_k, \bu_k,\bz_k),
& \quad & k = 0,\ldots, N \label{eq:DOCPpath} \\
& 0           & \geq & \; h_i(\bmu_k, \bu_k, \bz_k) + \be_i(\bmu_k, \bu_k, \bP_k, \bz_k),
& \quad & k = 1,\ldots, N;\; i \in \calI \label{eq:DOCPconstraints}
\end{alignat}
\end{subequations}
To perform the discretization, the track centerline is parameterized using a curvilinear parameter $\alpha \in [0,1]$, and uniformly sampled at $N+1$ points $\alpha_0, \ldots, \alpha_N$.
Accordingly, $\bmu_k$ denotes the mean state at grid node $\alpha_k$, while the controls $\bu_k$ and the algebraic variables $\bz_k$ are assumed to be piecewise constant over each interval $[\al_k, \al_{k+1}]$.
The $\bz_k$'s are introduced as direct handles for physically meaningful quantities such contact forces. Similarly, $\bP_k$ represents the covariance matrix at the $k$-th node.
Following the direct collocation approach, both the mean and covariance state trajectories are approximated, in the $k$-th interval, with polynomials $\pi_k(\tau)$, defined on the unit interval $\tau\in[0,1]$, and then scaled to match the width $\nu_k$ of the corresponding time step. On the unit interval, we select $d$ collocation points $\tau_1, \ldots, \tau_d$, associated with as many collocation states. Accordingly, we define $\bxi_k$ and $\bSi_k$ as the mean states and covariance matrices at the $d$ collocation points within each interval $[\al_k, \al_{k+1}]$, respectively. Therefore, $\bxi_k =
(\bxi_{k,1}, \ldots, \bxi_{k,d})$, and similarly $\bSi_k=(\bSi_{k,1}, \ldots, \bSi_{k,d})$. Equations~\eqref{eq:DOCPdyn} and~\eqref{eq:DOCPdP} represent the collocation and continuity equations for mean and covariance, respectively, with initial conditions represented by~\eqref{eq:DOCPdynIC} and~\eqref{eq:DOCPdPIC}. Eqs.~\eqref{eq:DOCPpath} are path equality constraints and~\eqref{eq:DOCPconstraints} are the robustified inequality constraints.

In certain cases~\cite{Gillis:PracticalMethodsApproximate:2015}, positive-definiteness-preserving Lyapunov discretization schemes are used in~\eqref{eq:DOCPdP}. However, for sufficiently fine discretization, we found that integrating only the lower triangular part of $\bP(t)$ in~\eqref{eq:OCPdPIC}  (and reconstructing its strictly upper part accordingly), ensures that $\bP_k$ remain symmetric and positive definite when the initial $\bP_0$ is symmetric and positive definite. In all our tests direct collocation with cubic polynomial state representations and Gauss-Legendre collocation points provided accurate results.

\subsection{Robust friction limit constraint formulation}
\label{sec:FLC}
The model employs a tire formulation that neglects combined slip effects, i.e., it does not account for the simultaneous utilization of longitudinal and lateral tire forces. As a result, the optimal control problem must include an additional constraint to ensure that the total ground reaction forces remain within the bounds of the tire's adherence ellipse.

The base version of the constraint for each axle can be expressed as the following inequality depending on $\bx$ and $\bu$
\begin{equation}
	h^\textrm{FLC}_j(\bx,\bu) =  S_j(\bx,\bu) - 1 \leq 0, \qquad{(j=1,2)}
\label{eq:adherence_with_axle_saturation}
\end{equation}
where we introduced the \emph{axle saturation ratio} $S_j(\bx,\bu)$ as follows
\begin{equation}
	S_j(\bx,\bu) = \frac{ \left( \frac{X_j(\bx,\bu)}{\mu_{x,j}} \right)^2+ \left( \frac{Y_j(\bx,\bu)}{\mu_{y,j}}\right)^2}{Z_j^2(\bx,\bu)}.
\label{eq:axle_saturation}
\end{equation}
In~\eqref{eq:adherence_with_axle_saturation}, the superscript FLC denotes friction limit constraint, and $j=1,2$ refer the front and rear axle, respectively. In~\eqref{eq:axle_saturation}, $X_j$ and $Y_j$ denote the longitudinal and lateral components of the in-plane ground forces, respectively, as illustrated in Fig.~\ref{fig:vehicle_model}, while $Z_j$ represents the total vertical load acting on the axle.

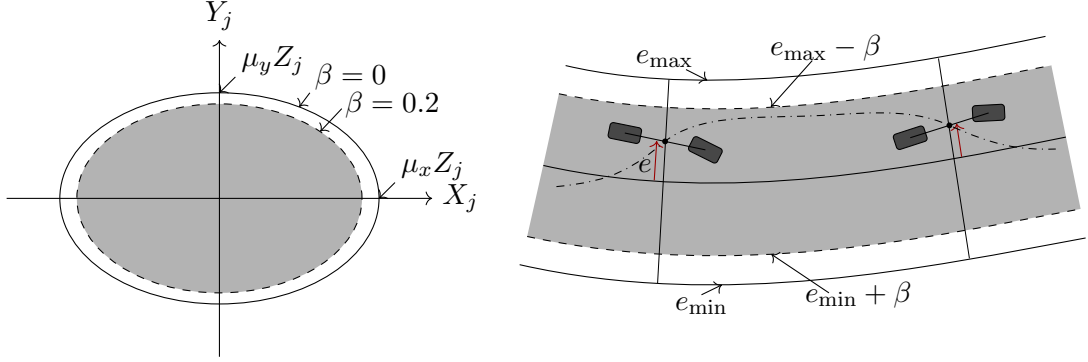
\begin{figure}
	\centering
	\adjustbox{valign=c}{\input{TikZ_Sources/tikz_adherence_constraint.tex}}
	\hfill
	\adjustbox{valign=c}{\input{TikZ_Sources/tikz_stayontrack_constraint.tex}}
	\caption{Graphical representation of the two constraints analyzed in this section: the friction limit constraint in the $X_jY_j$ plane (left panel), and the track limit constraint in the $xy$ plane (right panel). In both figures, the filled gray area bounded by a dashed line indicates the accessible region in the presence of a back-off term.
	}
	\label{fig:robust_constraints}
\end{figure}

From~\eqref{eq:axle_saturation} it is evident that $S_j(\bx,\bu)\in [0,1]$ indicates how close each configuration operates to the friction limit: $S_j=0$ when the overall grip demand is zero, while $S_j = 1$ when the point $\left(X_j, Y_j\right)$ lies exactly on the friction ellipse, i.e., under full saturation.
The points $\left(X_j,Y_j\right)$ are constrained to lie within an ellipse whose semi-axes are given by $\mu_{x,j}Z_j$ and $\mu_{y,j}Z_j$. The constraint, without the back-off term, is represented in the left panel of Fig.~\ref{fig:robust_constraints} by the solid line ellipse in the $X_jY_j$-plane.

To better clarify the effect of the back-off term on the constraint let us consider a combination of $\bx$, $\bu$, $P$ and $\ga^\textrm{FLC}_j$ so that $\be^\textrm{FLC}_j=0.2$. This implies that the semi-axes of the ellipse are reduced by a factor $\sqrt{1-\be^\textrm{FLC}_j} = \sqrt{0.8}$, as shown in the left panel of Fig.~\ref{fig:robust_constraints}, where the ellipse in dashed line delimits the available region with the back-off applied. This allows the optimizer to determine the most appropriate trade-off between the longitudinal force $X_j$ and the lateral force $Y_j$ --- often by reducing both components to some extent --- based on the specific requirements of the manoeuvre.


\subsection{Robust track limit constraint formulation}
\label{sec:TLC}
A track limit constraint is required to ensure that the vehicle remains within the circuit boundaries. The formulation adopted introduces an algebraic variable $e$, which satisfies the following constraint:
\begin{equation}
	\begin{Bmatrix}
		x\\y
	\end{Bmatrix}
	-\boldsymbol{c} - e\bn=\boldsymbol{0} \label{eq:onplane_constraint}
\end{equation}
where $\boldsymbol{c}$ indicates the position of the centerline, and $\bn$ is the unit normal vector to the centerline. Equation~\eqref{eq:onplane_constraint} constraints the CoM of the vehicle to lie on the vertical plane defined by the unit normal vector $\bn$. The previously introduced algebraic variable $e$ is further constrained to take values between $e_\textrm{min}$ and $e_\textrm{max}$, which are determined based on the width of the circuit and the vehicle's front and rear tracks.

In order to take into account the back-off terms, the maximum and minimum values are modified coherently, so that:
\begin{equation}
	e_\textrm{min} + \be^\textrm{TLC} \leq e \leq e_\textrm{max} - \be^\textrm{TLC}\label{eq:TLC}
\end{equation}
where $\be^\textrm{TLC}$ is computed by considering that the gradient of the constraint w.r.t. the state vector $\na_{\bx}h^\textrm{TLC}$ depends solely on $\bn$.

%% file: TikZ_Sources/tikz_vehicle_model.tex
\begin{tikzpicture}[scale=2, x = {(0,1cm)}, y = {(-1cm,0)},
	force/.style={->, thick, red!60!black, >={Latex[length=10pt, width=4pt]}},
	vec/.style={->, thick, mgreen, >={Latex[length=10pt, width=4pt]}},
	aux/.style={thin},
	frame/.style={->,thin,>={Latex}},
	every node/.style={font=\small, text=black}
	]
	\definecolor{mgreen}{RGB}{0,100,0}
	\begin{scope}[rotate = 25,shift = {(-.3,-2)}]
		\coordinate (rear) at (0,0);
		\coordinate (G) at (1.2,0);
		\coordinate (front) at (2.8,0);
		
		\draw[blue!50!black, thick] (rear) -- (front);
		
		\draw[fill=black!60!white, draw=black, rounded corners=2pt]
		($(rear)+(-.2,.1)$) rectangle ($(rear)+(+.2,-.1)$);
		
		\begin{scope}[shift={(front)}, rotate=25]
			\draw[fill=black!60!white, draw=black, rounded corners=2pt]
			(-0.2,.1) rectangle (0.2,-.1);
		\end{scope}
		
		\draw[force] (rear) -- ++(.5,0) node[above,right] {$X_2$};
		\draw[force] (rear) -- ++(0,1.3) node[above] {$Y_2$};
		
		\draw[force] (front) -- ++(-.3,0) node[right] {$X_1<0$};
		\draw[force] (front) -- ++(0,1.5) node[above] {$Y_1$};
		
		\draw[vec] (G) -- ++(.6,0) node[left, shift={(-.2,-.1)}] {$u\,\mathbf{i}$};
		\draw[vec] (G) -- ++(0,.3) node[above] {$v\,\mathbf{j}$};
		
		\begin{scope}[rotate = -115]
			\draw[vec] 
			(-.1,.8) arc[start angle=0, end angle=50, radius=1]
			node[midway, right=2pt] {$r\,\mathbf{k}$};
		\end{scope}
		
		\fill (G) circle (1pt) node[right] {$G$};
		
		\draw[frame] (G) --++ (1,0) node[right] {$x_b$};
		\draw[frame] (G) --++ (0,1) node[above] {$y_b$};
		
		
		\draw[aux] (front) -- ++(.4,0);
		\draw[aux] (front) -- ++({0.4*cos(25)}, {0.4*sin(25)});
		
		\draw[->,thin,>={Latex}] ([shift={(0.35,0)}]front) arc[start angle=0, end angle=25, radius=0.35];
		\node[left,shift={(.15,0)}] at ([shift={(0.35,0)}]front) {$\delta$};
		
		\draw[aux] (G) -- ++({0.4*cos(25)}, {-0.4*sin(25)});
		
		\draw[->,thin,>={Latex}] ([shift={({0.35*cos(25)},{-.35*sin(25)})}]G) arc[start angle=-25, end angle=0, radius=0.35];
		\node[left,shift={(.2,-.05)}] at ([shift={({0.35*cos(25)},{-.35*sin(25)})}]G) {$\psi$};
		
	\end{scope}
	\coordinate (ground) at (0,0);
	\draw[frame] (ground) --++ (1,0) node[right] {$x$};
	\draw[frame] (ground) --++ (0,1) node[above] {$y$};
	\draw[->,thick,>=Latex] (ground) --++ (G);
	
\end{tikzpicture}

%% file: TikZ_Sources/tikz_adherence_constraint.tex
\begin{tikzpicture}[scale = .7]
	\def\outerX{3}   
	\def\outerY{2}   
	\def\gap{0.2}    
	\def\innerX{2.68}
	\def\innerY{1.79}
	
	\draw[] (0,0) ellipse [x radius=\outerX, y radius=\outerY];
	
	\fill[black!30] (0,0) ellipse [
	x radius={\innerX},
	y radius={\innerY}
	];
	\draw[dashed] (0,0) ellipse [
	x radius={\innerX},
	y radius={\innerY}
	];
	
	\draw[->] (-\outerX-1,0) -- (\outerX+1,0) node[right] {$X_j$};
	\draw[->] (0,-\outerY-1) -- (0,\outerY+1) node[above] {$Y_j$};
	
	\draw[->,thin] (1.7,2) -- ({\outerX*cos(60)},{\outerY*sin(60)}) node[midway, above right] {$\beta=0$};
	\draw[->,thin] (2.3,1.7) -- ({(\innerX)*cos(45)},{(\innerY)*sin(45)}) node[midway, right, shift={(0.05,.2)}] {$\beta=\gap$};
	
	\draw[->,thin] ({\outerX+.3},.3) -- (\outerX,0) node[midway, above right,shift={(.05,0)}] {$\mu_x Z_j$};
	\draw[->,thin] (.3,{\outerY+.3}) -- (0,\outerY) node[midway, above right,shift={(.05,0)}] {$\mu_y Z_j$};
	
\end{tikzpicture}

%% file: TikZ_Sources/tikz_stayontrack_constraint.tex
\begin{tikzpicture}[scale = 1.1]
	\tikzset{sref/.style={
			draw = blue,
			->
	},
	elat/.style={
		red!60!black,
		->
	},
	elimits/.style={
	->
	},
	nplane/.style={
		very thin,
		draw=black
	}
}
\draw  [draw opacity=0][fill = black!30] (-1.430,1.400) .. controls (-1.295,1.353) and (0.079,1.126) .. (1.413,1.259) .. controls (2.446,1.339) and (3.866,1.619) .. (4.460,1.742) .. controls (4.513,1.493) and (4.737,0.363) .. (4.810,0.012) .. controls (3.786,-0.174) and (3.079,-0.348) .. (1.613,-0.501) .. controls (0.219,-0.621) and (-1.014,-0.508) .. (-1.810,-0.320) .. controls (-1.736,-0.005) and (-1.466,1.235) .. (-1.430,1.400) -- cycle ;
\draw    (-1.616,0.536) .. controls (0.594,0.038) and (2.894,0.528) .. (4.634,0.878) ;
\draw    (-1.350,1.740) .. controls (0.500,1.302) and (2.930,1.782) .. (4.390,2.092) ;
\draw[dashed]    (-1.430,1.400) .. controls (0.820,0.942) and (2.990,1.462) .. (4.460,1.742) ;
\draw    (-1.900,-0.660) .. controls (0.550,-1.218) and (2.990,-0.688) .. (4.870,-0.338) ;
\draw[dashed]    (-1.810,-0.320) .. controls (0.700,-0.858) and (2.930,-0.338) .. (4.810,0.012) ;
\draw  [nplane]  (-0.114,1.573) -- (-0.250,-0.884) ;
\draw  [nplane]  (3.110,1.826) -- (3.480,-0.584) ;

\draw [elat]   (-0.290,0.357) -- (-0.260,0.832) node[midway,black,xshift=-4,yshift=-2] {$e$};;
\draw[elat]   (3.387,0.641) -- (3.326,1.042) ;

\fill (-0.160,0.826) circle (1pt) ;
\fill (3.241,1.019) circle (1pt) ;
\draw[dashdotted]    (-1.460,0.281) .. controls (-0.767,0.347) and (-0.540,0.514) .. (-0.160,0.826) .. controls (0.220,1.137) and (0.886,1.113) .. (1.496,1.130) .. controls (2.107,1.147) and (2.715,1.258) .. (3.241,1.019) .. controls (3.767,0.781) and (4.053,0.721) .. (4.513,0.734) ;
\draw    (-0.642,0.927) -- (0.322,0.724) ;
\draw  [fill={rgb, 255:red, 0; green, 0; blue, 0 }  ,fill opacity=0.6 ] (0.176,0.862) .. controls (0.186,0.881) and (0.209,0.888) .. (0.228,0.879) -- (0.501,0.742) .. controls (0.521,0.733) and (0.528,0.709) .. (0.519,0.690) -- (0.467,0.587) .. controls (0.457,0.568) and (0.434,0.560) .. (0.415,0.569) -- (0.142,0.706) .. controls (0.122,0.716) and (0.115,0.739) .. (0.124,0.758) -- cycle ;
\draw  [fill={rgb, 255:red, 0; green, 0; blue, 0 }  ,fill opacity=0.6 ] (-0.818,1.022) .. controls (-0.814,1.042) and (-0.793,1.056) .. (-0.772,1.052) -- (-0.473,0.991) .. controls (-0.452,0.987) and (-0.438,0.967) .. (-0.442,0.946) -- (-0.465,0.833) .. controls (-0.469,0.812) and (-0.490,0.798) .. (-0.511,0.802) -- (-0.810,0.863) .. controls (-0.831,0.867) and (-0.845,0.887) .. (-0.841,0.908) -- cycle ;
\draw    (2.772,0.870) -- (3.710,1.169) ;
\draw  [fill={rgb, 255:red, 0; green, 0; blue, 0 }  ,fill opacity=0.6 ] (3.515,1.217) .. controls (3.514,1.238) and (3.531,1.256) .. (3.552,1.257) -- (3.857,1.273) .. controls (3.879,1.274) and (3.897,1.258) .. (3.898,1.237) -- (3.904,1.121) .. controls (3.905,1.100) and (3.889,1.082) .. (3.868,1.081) -- (3.562,1.065) .. controls (3.541,1.063) and (3.523,1.080) .. (3.521,1.101) -- cycle ;
\draw  [fill={rgb, 255:red, 0; green, 0; blue, 0 }  ,fill opacity=0.6 ] (2.572,0.865) .. controls (2.565,0.886) and (2.576,0.907) .. (2.597,0.914) -- (2.887,1.009) .. controls (2.908,1.016) and (2.929,1.005) .. (2.936,0.985) -- (2.972,0.875) .. controls (2.979,0.854) and (2.968,0.833) .. (2.948,0.826) -- (2.657,0.731) .. controls (2.637,0.724) and (2.615,0.735) .. (2.608,0.755) -- cycle ;

\draw[elimits]    (0.090,1.692) -- (0.345,1.562) node[above,xshift = -16,yshift = 1] {$e_\textrm{max}$};
\draw[elimits]   (0.275,-1.023) node[below,yshift = 3] {$e_\textrm{min}$} -- (0.520,-0.903) ;
\draw[elimits]     (1.6,1.8) node[above,yshift = -4,xshift = 5] {$e_\textrm{max}-\be$} -- (1.105,1.242) ;
\draw[elimits]     (1.8,-1.0) node[below,yshift = 8,xshift = 11] {$e_\textrm{min}+\be$} -- (1.110,-0.534) ;
	
\end{tikzpicture}

%% file: open_loop_planning.tex
\section{Open-loop planning via H-steps ahead predictions}
\label{sec:open_loop_planning}
In this section, we address the problem of planning under uncertainty for a stochastic system.
At each grid point $k$ of the discretized trajectory the covariance matrix is propagated in a feed-forward fashion for a fixed prediction horizon of $H$ steps.
In the planning problem, we conservatively evaluate the maximum propagation of the covariance, that is, the matrix evolved over $H$ steps, to ensure the tightest possible enforcement of constraints. This maximizes the robustness of the resulting solution, as the constraint back-off is evaluated in the worst-case scenario within the prediction horizon.

To enable this, we associate multiple versions of the covariance matrix with each grid point. Specifically, $H+1$ instances of the matrix $\bP$ are maintained at each step. To this sake we introduce the notation $\bP_k^j$, where $j = 0, \ldots, H$, and $\bP_k^j$ represents the version of the covariance matrix at step $k$ that was initialized $j$ steps earlier. Hence, $\bP_k^0$ originates at node $k$ itself and will be propagated forward and employed in subsequent steps (exactly when evolved as $\bP_{k+H}^H$ ), while $\bP_k^H$ is the instance that originated $H$ steps earlier and is the most propagated one at step $k$.

The repeated initialization of the covariance matrix at each step is necessary to prevent its unbounded growth, as we assume that the system evolves without feedback control.
In the absence of feedback, the uncertainty on the position and orientation of the body frame accumulates along the track, leading to a severe and unrealistic degradation of the covariance matrix conditioning.
For this reason, the propagation horizon in this open-loop setting is a critical design parameter.
We bound the number of steps $H$ to realistically capture the evolution of disturbances occurring prior to any corrective driver response.

This approach requires modifications to the formulation~\eqref{eq:DOCP} introduced in Sec.~\ref{sec:discretization}. In particular, Eqs.~\eqref{eq:DOCPdP}, \eqref{eq:DOCPdPIC}, and \eqref{eq:DOCPconstraints} are replaced by Eqs.~\eqref{eq:DOCPdPopenloop}, \eqref{eq:DOCPdPinitopenloop}, and \eqref{eq:DOCPconstraintsopenloop}. Equation~\eqref{eq:DOCPdPopenloop} collects the continuity and collocation equations for all versions of the covariance matrix propagated from previous steps. To support the evolution of $H$ versions, we introduce $H \times d$ matrices $\bSi_k^j = (\bSi_{k,1}^j, \ldots, \bSi_{k,d}^j)$, representing the collocation values of each version $j$ of the covariance matrix within interval $k$.

%

The specific formulation is expressed as follows:
\begin{subequations}\label{eq:DOCPopenloop}
\begin{alignat}{3}
	\underset{\bmu_k,\bxi_k, \bu_k, \bP_k,\bz_k}{\text{minimize}} \,
	& & & J_k(\bmu_k,\bxi_k, \bu_k) & & \label{eq:DOCPcostopenloop} \\
	\hspace*{-2.0 cm}\text{s.t.} \quad
	& \bzero      & = & \; \bPsimu_k(\bmu_{k-1},\bmu_k,\bxi_k, \bu_k,\bz_k),
	& \quad & k = 1,\ldots, N \label{eq:DOCPdynopenloop} \\
	& \bmu_0      & = & \; \bar{\bmu}_0
	& & \label{eq:DOCPdynICopenloop} \\
	& \bzero      & = & \; \bPsiP_k(\bmu_k,\bxi_k, \bu_k, \bP_{k-1}^{j-1},\bP_k^j,\bSi_k^j,\bz_k),
	& \quad & \substack{k = 1,\ldots, N; \; \\ j=1,\ldots,H;\;} \label{eq:DOCPdPopenloop} \\
	& \bP_k^0       & = & \; \bar{\bP}_0 \succeq 0
	& \quad & k = 1,\ldots, N; \; \label{eq:DOCPdPinitopenloop} \\
	& \bzero      & = & \; \bOm_k(\bmu_k,\bxi_k, \bu_k,\bz_k),
	& \quad & k = 0,\ldots, N \label{eq:DOCPpathopenloop} \\
	& 0           & \geq & \; h_i(\bmu_k, \bu_k, \bz_k) + \be_i(\bmu_k, \bu_k, \bP_k^H, \bz_k),
	& \quad & k = 1,\ldots, N;\; i \in \calI \label{eq:DOCPconstraintsopenloop}
\end{alignat}
\end{subequations}
Equation~\eqref{eq:DOCPdPinitopenloop} defines the initialization of the appropriate covariance matrix version at step $k$, while Eq.~\eqref{eq:DOCPconstraintsopenloop} specifies that the back-off term in the constraints is computed using the most propagated version, $\bP^H_k$. A schematic illustration of the management of the $H+1$ covariance matrix instances is provided in Fig.~\ref{fig:DOCPgrid}. The dashed rectangle highlights the $k$-th discretization step, where the constraints are evaluated. At each grid point, two particular versions of the covariance matrix are emphasized: $\bP^0_k$ (red node), representing the instance to be initialized at that step, and $\bP^H_k$ (green node), corresponding to the version that has been propagated over the full horizon of $H$ steps and is used to robustify the constraints in Eq.~\eqref{eq:DOCPconstraintsopenloop}.

\def\version{marco}
\ifdefstring{\version}{matteo}{
\begin{figure}
	\centering
	\input{TikZ_Sources/tikz_covariance_evolution_matteo.tex}
	\caption{Schematic representation of the continuity and initialization equations associated with the $H+1$ propagated instances of the covariance matrix introduced in problem~\eqref{eq:DOCPopenloop}. The dashed rectangle highlights the $k$-th discretization step, at which the robust constraints are enforced. Each node corresponds to a specific covariance matrix, where the subscript denotes the current grid index and the superscript indicates the number of propagation steps since initialization. At each step, two key instances are emphasized: the matrix to be initialized at that step (red node), and the matrix that has been propagated over $H$ steps (green node), which is used in the evaluation of the constraint back-off.}
	\label{fig:DOCPgrid}
\end{figure}
}{}

\ifdefstring{\version}{marco}{
\begin{figure}
	\centering
	\definecolor{mygreen}{RGB}{0,150,0}
	\input{TikZ_Sources/tikz_covariance_evolution.tex}
	\caption{Schematic representation (depicted for $H=3$) of the continuity and initialization equations associated with the $H+1$ propagated instances of the covariance matrix introduced in problem~\eqref{eq:DOCPopenloop}. The dashed rectangle highlights the $k$-th discretization step, at which the robust constraints are enforced. Each node corresponds to a specific covariance matrix, where the subscript denotes the current grid index and the superscript indicates the number of propagation steps since initialization. At each step, two key instances are emphasized: the matrix to be initialized at that step (red node), and the matrix that has been propagated over $H$ steps (green node), which is used in the evaluation of the constraint back-off.}
	\label{fig:DOCPgrid}
\end{figure}
}{}

%% file: TikZ_Sources/tikz_covariance_evolution_matteo.tex
\begin{tikzpicture}[%
	smallnode/.style={%
		circle, draw, minimum size=3mm, inner sep=0pt
	},
	r_smallnode/.style={%
		circle, draw, minimum size=3mm, inner sep=0pt, red
	},
	g_smallnode/.style={%
		circle, draw, minimum size=3mm, inner sep=0pt, mygreen
	},
	every path/.style={->, thick},
	x=1.5cm, y=1.5cm
	]
	
	\newcommand{\Plab}[2]{\bP^{#2}_{#1}}
	\definecolor{mygreen}{RGB}{0 150 0}
	
	\node[smallnode, label={[label distance=-2pt]above:\(\Plab{k-2}{H-2}\)}] (Pk-2H-2) at (1,0) {};
	\node[smallnode, label={[label distance=-2pt]above:\(\Plab{k-1}{H-1}\)}] (Pk-1H-1) at (2,0) {};
	\node[g_smallnode, label={[label distance=-2pt]above:\(\Plab{k}{H}\)}] (PkH) at (3,0) {};
	\node[r_smallnode, label={[label distance=-2pt]above:\(\Plab{k+1}{0}\)}] (Pk+10) at (4,0) {};
	\node[smallnode, label={[label distance=-2pt]above:\(\Plab{k+2}{1}\)}] (Pk+21) at (5,0) {};
	
	\draw[bend left=20] (Pk-2H-2) to (Pk-1H-1);
	\draw[bend left=20] (Pk-1H-1) to (PkH);
	\draw[bend left=20] (Pk+10) to (Pk+21);
	
	\node[r_smallnode, label={[label distance=-2pt]above:\(\Plab{k-H}{0}\)}] (Pk-H0) at (0,0) {};
	\node[label={center:\(\dots\)}] (dots10) at (-.5, 0) {};
	\node[label={center:\(\dots\)}] (dots11) at (.5, 0) {};
	\node[smallnode, label={[label distance=-2pt]above:\(\Plab{k+H}{H-1}\)}] (Pk+HH-1) at (6,0) {};
	\node[label={center:\(\dots\)}] (dots12) at (5.5, 0) {};
	\node[label={center:\(\dots\)}] (dots13) at (6.5, 0) {};
	\node[smallnode, label={[label distance=-2pt]above:\(\Plab{k-2}{H-3}\)}] (Pk-2H-3) at (1,1) {};
	\node[smallnode, label={[label distance=-2pt]above:\(\Plab{k-1}{H-2}\)}] (Pk-1H-2) at (2,1) {};
	\node[smallnode, label={[label distance=-2pt]above:\(\Plab{k}{H-1}\)}] (PkH-1) at (3,1) {};
	\node[g_smallnode, label={[label distance=-2pt]above:\(\Plab{k+1}{H}\)}] (Pk+1H) at (4,1) {};
	\node[r_smallnode, label={[label distance=-2pt]above:\(\Plab{k+2}{0}\)}] (Pk+20) at (5,1) {};
	
	\draw[bend left=20] (Pk-2H-3) to (Pk-1H-2);
	\draw[bend left=20] (Pk-1H-2) to (PkH-1);
	\draw[bend left=20] (PkH-1) to (Pk+1H);
	
	\node[g_smallnode, label={[label distance=-2pt]above:\(\Plab{k-H}{H}\)}] (Pk-HH) at (0,1) {};
	\node[label={center:\(\dots\)}] (dots20) at (-.5, 1) {};
	\node[label={center:\(\dots\)}] (dots21) at (.5, 1) {};
	\node[smallnode, label={[label distance=-2pt]above:\(\Plab{k+H}{H-2}\)}] (Pk+HH-2) at (6,1) {};
	\node[label={center:\(\dots\)}] (dots22) at (5.5, 1) {};
	\node[label={center:\(\dots\)}] (dots23) at (6.5, 1) {};
	
	\node[label={center,rotate=90:\(\dots\)}] (dots1) at (0, 1.7) {};
	\node[label={center,rotate=90:\(\dots\)}] (dots2) at (1, 1.7) {};
	\node[label={center,rotate=90:\(\dots\)}] (dots3) at (2, 1.7) {};
	\node[label={center,rotate=90:\(\dots\)}] (dots4) at (3, 1.7) {};
	\node[label={center,rotate=90:\(\dots\)}] (dots5) at (4, 1.7) {};
	\node[label={center,rotate=90:\(\dots\)}] (dots6) at (5, 1.7) {};
	\node[label={center,rotate=90:\(\dots\)}] (dots7) at (6, 1.7) {};
	
	\node[g_smallnode, label={[label distance=-2pt]above:\(\Plab{k-2}{H}\)}] (Pk-2H) at (1,2) {};
	\node[r_smallnode, label={[label distance=-2pt]above:\(\Plab{k-1}{0}\)}] (Pk-10) at (2,2) {};
	\node[smallnode, label={[label distance=-2pt]above:\(\Plab{k}{1}\)}] (Pk1) at (3,2) {};
	\node[smallnode, label={[label distance=-2pt]above:\(\Plab{k+1}{2}\)}] (Pk+12) at (4,2) {};
	\node[smallnode, label={[label distance=-2pt]above:\(\Plab{k+2}{3}\)}] (Pk+23) at (5,2) {};
	
	\draw[bend left=20] (Pk-10) to (Pk1);
	\draw[bend left=20] (Pk1) to (Pk+12);
	\draw[bend left=20] (Pk+12) to (Pk+23);
	
	\node[smallnode, label={[label distance=-2pt]above:\(\Plab{k-H}{2}\)}] (Pk-H2) at (0,2) {};
	\node[label={center:\(\dots\)}] (dots30) at (-.5, 2) {};
	\node[label={center:\(\dots\)}] (dots31) at (.5, 2) {};
	\node[r_smallnode, label={[label distance=-2pt]above:\(\Plab{k+H}{0}\)}] (Pk+H0) at (6,2) {};
	\node[label={center:\(\dots\)}] (dots32) at (5.5, 2) {};
	\node[label={center:\(\dots\)}] (dots33) at (6.5, 2) {};
	
	\node[smallnode, label={[label distance=-2pt]above:\(\Plab{k-2}{H-1}\)}] (Pk-2H-1) at (1,3) {};
	\node[g_smallnode, label={[label distance=-2pt]above:\(\Plab{k-1}{H}\)}] (Pk-1H) at (2,3) {};
	\node[r_smallnode, label={[label distance=-2pt]above:\(\Plab{k}{0}\)}] (Pk0) at (3,3) {};
	\node[smallnode, label={[label distance=-2pt]above:\(\Plab{k+1}{1}\)}] (Pk+11) at (4,3) {};
	\node[smallnode, label={[label distance=-2pt]above:\(\Plab{k+2}{2}\)}] (Pk+22) at (5,3) {};
	
	\draw[bend left=20] (Pk-2H-1) to (Pk-1H);
	\draw[bend left=20] (Pk0) to (Pk+11);
	\draw[bend left=20] (Pk+11) to (Pk+22);
	
	\node[smallnode, label={[label distance=-2pt]above:\(\Plab{k-H}{1}\)}] (Pk-H1) at (0,3) {};
	\node[label={center:\(\dots\)}] (dots40) at (-.5, 3) {};
	\node[label={center:\(\dots\)}] (dots41) at (.5, 3) {};
	\node[g_smallnode, label={[label distance=-2pt]above:\(\Plab{k+H}{H}\)}] (Pk+HH) at (6,3) {};
	\node[label={center:\(\dots\)}] (dots42) at (5.5, 3) {};
	\node[label={center:\(\dots\)}] (dots43) at (6.5, 3) {};
	
	\draw[dashed, thick, rounded corners] (2.6,-.25) rectangle (3.4,3.5);
	
\end{tikzpicture}

%% file: TikZ_Sources/tikz_covariance_evolution.tex
\begin{tikzpicture}[
	smallnode/.style={circle, draw, minimum size=3mm, inner sep=0pt},
	r_smallnode/.style={circle, draw=red, minimum size=3mm, inner sep=0pt},
	g_smallnode/.style={circle, draw=mygreen, minimum size=3mm, inner sep=0pt},
	r_hatnode/.style={
		circle, draw=red, pattern=north east lines, pattern color=red,
		minimum size=3mm, inner sep=0pt},
	g_hatnode/.style={
		circle, draw=mygreen, pattern=north east lines, pattern color=mygreen,
		minimum size=3mm, inner sep=0pt},
	every path/.style={->, thick},
	x=1.4cm, y=1.2cm
	]
	
	\node[r_hatnode, label={[label distance=-2pt]above:\(\bP_{k-3}^0\)}] (P00) at (1,3) {};
	\node[smallnode, label={[label distance=-2pt]above:\(\bP_{k-2}^1\)}] (P11) at (2,3) {};
	\node[smallnode, label={[label distance=-2pt]above:\(\bP_{k-1}^2\)}] (P22) at (3,3) {};
	\node[g_hatnode, label={[label distance=-2pt]above:\(\bP_k^3\)}] (P33) at (4,3) {};
	\node[r_hatnode, label={[label distance=-2pt]above:\(\bP_{k+1}^0\)}] (P44) at (5,3) {};
	\node[smallnode, label={[label distance=-2pt]above:\(\bP_{k+2}^1\)}] (P55) at (6,3) {};
	
	\node[r_hatnode, label={[label distance=-2pt]above:\(\bP_{k-2}^0\)}] (P10) at (2,2) {};
	\node[smallnode, label={[label distance=-2pt]above:\(\bP_{k-1}^1\)}] (P21) at (3,2) {};
	\node[smallnode, label={[label distance=-2pt]above:\(\bP_k^2\)}] (P32) at (4,2) {};
	\node[g_hatnode, label={[label distance=-2pt]above:\(\bP_{k+1}^3\)}] (P43) at (5,2) {};
	\node[r_hatnode, label={[label distance=-2pt]above:\(\bP_{k+2}^0\)}] (P54) at (6,2) {};
	\node[smallnode, label={[label distance=-2pt]above:\(\bP_{k+3}^1\)}] (P65) at (7,2) {};

	\node[r_hatnode, label={[label distance=-2pt]above:\(\bP_{k-1}^0\)}] (P20) at (3,1) {};
	\node[smallnode, label={[label distance=-2pt]above:\(\bP_k^1\)}] (P31) at (4,1) {};
	\node[smallnode, label={[label distance=-2pt]above:\(\bP_{k+1}^2\)}] (P42) at (5,1) {};
	\node[g_hatnode, label={[label distance=-2pt]above:\(\bP_{k+2}^3\)}] (P53) at (6,1) {};
	\node[r_hatnode, label={[label distance=-2pt]above:\(\bP_{k+3}^0\)}] (P64) at (7,1) {};
	\node[smallnode, label={[label distance=-2pt]above:\(\bP_{k+4}^1\)}] (P75) at (8,1) {};
	
	\node[r_hatnode, label={[label distance=-2pt]above:\(\bP_k^0\)}] (P30) at (4,0) {};
	\node[smallnode, label={[label distance=-2pt]above:\(\bP_{k+1}^1\)}] (P41) at (5,0) {};
	\node[smallnode, label={[label distance=-2pt]above:\(\bP_{k+2}^2\)}] (P52) at (6,0) {};
	\node[g_hatnode, label={[label distance=-2pt]above:\(\bP_{k+3}^3\)}] (P63) at (7,0) {};
	\node[r_hatnode, label={[label distance=-2pt]above:\(\bP_{k+4}^0\)}] (P74) at (8,0) {};
	\node[smallnode, label={[label distance=-2pt]above:\(\bP_{k+5}^1\)}] (P85) at (9,0) {};

	\draw (P00) -- (P11);
	\draw (P11) -- (P22);
	\draw (P22) -- (P33);
	\draw (P22) -- (P33);
	
	\draw (P44) -- (P55);
	\draw[dashed,->] (P55) -- ++(1,0);
	
	\draw (P10) -- (P21);
	\draw (P21) -- (P32);
	\draw (P32) -- (P43);
	\draw (P54) -- (P65);
	\draw[dashed,->] (P65) -- ++(1,0);

	\draw (P20) -- (P31);
	\draw (P31) -- (P42);
	\draw (P42) -- (P53);
	\draw (P64) -- (P75);
	\draw[dashed,->] (P75) -- ++(1,0);
	
	\draw (P30) -- (P41);
	\draw (P41) -- (P52);
	\draw (P52) -- (P63);
	
	\draw (P74) -- (P85);
	\draw[dashed,->] (P85) -- ++(1,0);
	
	\draw[dashed, thick, rounded corners] (3.6,-0.5) rectangle (4.4,3.7);
	
	\node at (0, 4.3) {\scriptsize \(\cdots\)};
	\node at (1, 4.3) {\scriptsize \(k-3\)};
	\node at (2, 4.3) {\scriptsize \(k-2\)};
	\node at (3, 4.3) {\scriptsize \(k-1\)};
	\node (knode) at (4, 4.3) {\scriptsize \(k\)};
	\node at (5, 4.3) {\scriptsize \(k+1\)};
	\node at (6, 4.3) {\scriptsize \(k+2\)};
	\node at (7, 4.3) {\scriptsize \(k+3\)};
	\node at (8, 4.3) {\scriptsize \(k+4\)};
	\node at (9, 4.3) {\scriptsize \(\cdots\)};
	
	\draw[->, thick, decorate, decoration={snake, amplitude=0.25mm, segment length=2.5mm}]
	(knode.south) -- (4, 3.8);
	
	%
	
\end{tikzpicture}

%% file: closed_loop_planning.tex
\section{Closed-loop uncertainty-aware planning over the full horizon}
\label{sec:closed_loop_planning}

The second approach we propose plans over the full horizon while incorporating a feedback policy that mimics the driver's closed-loop behavior. This enables a more realistic propagation of the uncertainty encoded in the covariance matrix $\bP$, preventing its uncontrolled growth through the stabilizing effect of the driver's feedback.

The approach is based on the three following steps. 
\begin{enumerate}[label=\roman*)]
\item A nominal time-optimal trajectory is planned: this will be referred to as nominal feed-forward optimal trajectory. This represents the trajectory that an expert driver would follow in an ideal scenario without uncertainty. In this step, the mean trajectory coincides with the actual path that the vehicle would follow under deterministic dynamics.

\item A closed-loop controller mimicking the driver's action is computed to stabilize the nominal feed-forward optimal trajectory obtained at step (i). We assume a discrete time-varying Linear Quadratic Regulator (LQR) controller is a good approximation of how an expert driver can track a prescribed trajectory.

\item In the last step the stochastic framework is reintroduced and the time-optimal planning problem is robustified using the Lyapunov framework. The main idea is to \emph{re-plan} a mean trajectory of a stochastic system such that, a time-varying LQR controller that aims at stabilizing it, is able to properly tame the propagation of the uncertainty. This amounts to ensuring that the robustified path constraints are satisfied. More specifically, this considers two key aspects: firstly, that the covariance matrix is propagated accounting explicitly for the controlled system dynamics; secondly, that the path constraints are robustly satisfied by incorporating the covariance estimation propagated through the closed-loop controlled system. Some aspects that may appear a bit technical are in fact essential. In this third step, not only are states and controls re-planned, but also the time-varying LQR gains. As a matter of fact, they need to explicitly account for the changed requirements associated with stabilizing a trajectory different from the nominal one, which must now satisfy the robustified constraints. 
\end{enumerate}

\subsection{Step 1 -- Nominal feed-forward time-optimal trajectory planning}
\label{sec:nominalFF}
Step (i) consists of planning a nominal feed-forward trajectory. This assumes deterministic system dynamics, where $\hbmu_k$, $\hbu_k$, and $\hbz_k$ denote differential states, controls, and algebraic states, respectively, for \( k = 0, \ldots, N \). To avoid redundancy, we note that the NLP to be solved is a special case of~\eqref{eq:DOCP}, where~\eqref{eq:DOCPdP},~\eqref{eq:DOCPdPIC}, and the back-off terms are omitted, since the goal is to compute a nominal trajectory.

\subsection{Step 2 -- Time-varying LQR stabilizing controller}
\label{sec:LQR}

Given the nominal trajectory $(\hbmu_k, \hbu_k)$ over $k=0,\ldots, N$, planned at step (i), and introduced the deviation variables $\bbmu_k=\bmu_k - \hbmu_k$ and $\bbu_k = \bu_k - \hbu_k$, we design a linear state-feedback law $\bbu_k= -\hbK_k \bbmu_k$, or equivalently $\bu_k = \hbu_k - \hbK_k (\bmu_k - \hbmu_k)$, to mitigate disturbances while tracking the nominal trajectory.

To compute the state-feedback matrices $\hbK_k$, we linearize the system around $(\hbmu_k, \hbu_k)$, such that $\bbmu_{k+1} \approx \hbA_{k}\bbmu_k + \hbB_k \bbu_k$, where $\hbA_k = f_{,\bmu}(\hbmu_k,\hbu_k)$ and $\hbB_k = f_{,\bu}(\hbmu_k,\hbu_k)$, and define the quadratic regulator (tracking) cost function
\begin{align}
J = \bbmu_N^T \bW_N \bbmu_N + \sum_{k=0}^{N-1} \big( \bbmu_k^{T} \bW_k \bbmu_k + \bbu_k^T \bR_k \bbu_k \big),
\end{align}
where $\bW_N=\bW_N^T \succ 0$, $\bW_k=\bW_k^T \succeq 0$, $\bR_k=\bR_k^T \succ 0$. It can be shown that the optimal cost-to-go is given by $J_k^*(\bbmu_k) = \bbmu_k^T \bS_k \bbmu_k$, with $\bS_k = \bS_k^T \succ 0$, where $\bS_k$ can be computed through the (backward) Riccati recursion for $k=N-1,\ldots, 0$
\begin{align}
\bS_N &= \bW_N,\\
\bS_{k} &= \bW_k + \hbA_k^T \bS_{k+1} \hbA_k - \hbA_k^T \bS_{k+1}\hbB_k \hbK_k,\quad k={N-1, \ldots, 0}
\end{align}
where  $\hbK_k =\big( \bR_k+\hbB_k^T \bS_{k+1} \hbB_k \big)^{-1} \hbB_k^T \bS_{k+1} \hbA_k$. The optimal (stabilizing) feedback policy is given by $\bbu_k=-\hbK_k \bbmu_k$. We assume that the gain sequence $\{\hbK_k\}$, $k=0,\ldots, N$ represents a reasonable approximation of the control strategy that an expert driver would adopt to optimally stabilize the system along the nominal trajectory.

\subsection{Step 3 -- Closed-loop robustified planning}
\label{sec:CLrobustified}
As the third and last step, a closed-loop robustified planning problem is set up. Its aim is the definition of a mean trajectory of a stochastic system which, working in tandem with a state-feedback controller that aims at stabilizing it, is able to mitigate the perturbations and reduce the propagation of uncertainty. This helps to guarantee that the robustified path constraints can be satisfied over the entire planning horizon. The discretized version of this problem takes the form of the following NLP:
\begin{subequations}\label{eq:ROCP}
\begin{alignat}{3}
\underset{\bmu_k,\bxi_k, \bu_k, \bP_k,\bz_k, \bK_k}{\text{minimize}}
    & \quad J_k(\bmu_k,\bxi_k, \bu_k) & & \label{eq:ROCPcost} \\
\text{s.t.}\notag\\
\bzero      & = \; \bPsimu_k(\bmu_{k-1},\bmu_k,\bxi_k, \bu_k,\bz_k),
\quad k = 1,\ldots, N \label{eq:ROCPdyn} \\
\bmu_0      & = \; \bar{\bmu}_0 \label{eq:ROCPdynIC} \\
\bzero      & = \; \bPsiPCL_k(\bmu_k,\bxi_k, \bu_k, \tbP_{k-1},\tbP_k,\bSi_k,\bz_k, \bK_k),
\quad k = 1,\ldots, N \label{eq:ROCPdP} \\
\tbP_0       & = \; \bar{\bP}_0 \succeq 0 \label{eq:ROCPdPIC} \\
\bzero      & = \; \bOm_k(\bmu_k,\bxi_k, \bu_k,\bz_k), \quad k = 0,\ldots, N \label{eq:ROCPpath} \\
0    & \geq  h_i(\bmu_k, \bu_k, \bz_k) + \be_i(\bmu_k, \bu_k, \tbP_k, \bz_k), \quad k = 1,\ldots, N;\; \quad i \in \calI \label{eq:ROCPconstraints}\\
-\de \bK_k   & \leq \bK_k - \hbK_k \leq  \de \bK_k \label{eq:ROCPdeltaK}, \quad k = 0,\ldots, N
\end{alignat}
\end{subequations}
Here,~\eqref{eq:ROCPdP} denotes the discrete-time covariance dynamics under the stabilizing state-feedback controller \( \bbu_k = -\bK_k \bbmu_k \). To distinguish this case from~\eqref{eq:DOCP}, we use the symbol \( \tbP \) to denote the closed-loop (CL) covariance matrix, whose discrete evolution corresponds to a collocation-based approximation of the continuous-time dynamics $\dot{\tbP}(t) = \tbA(t) \tbP(t) + \tbP(t)\tbA(t)^T + \bQ(t)$, with $\tbA(t)=\bA(t)-\bB(t)\bK(t)$. It is worth noting that, accordingly, in equation~\eqref{eq:ROCPconstraints} the back-off term $\be_i$ is a function of closed-loop uncertainty $\tbP_k$. Another key aspect is that the state-feedback matrix $\bK_k$ is also treated as a decision variable. This reflects the fact that the driver may adjust his action to stabilize the new (to-be-planned) nominal trajectory using a policy that differs from the previous $\hbK_k$, computed in step (ii) via LQR. To promote convergence, however, $\bK_k$ is restrained within prescribed bounds ($\de \bK_k$) relative to $\hbK_k$. In our tests $\de \bK_k = 0.1 \hbK_k$.

\subsection{Handling mutually exclusive throttle and braking commands}
In our formulation, the state vector is defined as $\bx=\big[u\;v\;r\;x_{G}\;y_{G}\;\psi \big]^T\in\bbR^6$, while the control input vector is $\bu=\big[F_x^a\;F_x^b\;\de \big]^T\in\bbR^3$. The longitudinal acceleration and braking forces, $F_x^a \geq 0$ and $F_x^b \leq 0$, respectively, are not independent but are subject to the \emph{complementarity constraint} $F_x^a F_x^b = 0$, which enforces mutual exclusivity between throttle and braking at any instant.

For convenience, we denote $u_1 = F_x^a \geq 0$ and $u_2 = F_x^b \leq 0$, so that the complementarity condition becomes $u_1 u_2 = 0$. Beyond the modeling constraint itself, this condition also imposes structural implications on admissible control variations, particularly relevant in the closed-loop formulation. 
Specifically, in a nominal \emph{acceleration} phase ($\hat{u}_1 > 0, \hat{u}_2 = 0$), both \(\delta u_1 \lesseqgtr 0\) are permitted, but only variations $\delta u_2 < 0$ are physically meaningful -- since a positive brake increment ($\delta u_2 > 0$) would unphysically contribute to propulsion. Conversely, in a \emph{braking} phase ($\hat{u}_2 < 0$, $\hat{u}_1 = 0$), $\delta u_2 \lesseqgtr 0$ are admissible, but only $\delta u_1 > 0$ should be allowed, as negative throttle increments ($\delta u_1 < 0$) would again introduce non-physical behavior.

To ensure physical consistency while preserving a tractable control structure, we introduce a \emph{minimal control vector} \(\tilde{\bu} = [F_x\;\; \delta]^T \in \mathbb{R}^2\), where \(F_x\) represents the net longitudinal force. The full control vector $\bu$ is then recovered via a state-dependent linear transformation $
\bu = \bH(\tilde{\bu}) \tilde{\bu}$ with
\begin{equation} \label{eq:minimalcontrol}
\bH(\tilde{\bu}) =
\begin{bmatrix}
\Ipos{F_x} & 0 \\
\Ineg{F_x} & 0 \\
0 & 1
\end{bmatrix}, \quad
\Ipos{x} :=
\begin{cases}
1, & \text{if } x \geq 0 \\
0, & \text{if } x < 0
\end{cases}, \quad
\Ineg{x} :=
\begin{cases}
0, & \text{if } x \geq 0 \\
1, & \text{if } x < 0
\end{cases}.
\end{equation}

In this formulation, \(\Ipos{x}\) and \(\Ineg{x}\) denote the \emph{positive-} and \emph{negative-part indicator functions}, respectively. Importantly, control variations also respect this mapping: $\delta\bu = \bH(\tilde{\bu}) \delta\tilde{\bu}$. Consequently, the control input matrix can be expressed as \(\bB \bH = \tilde{\bB}\), and the state-feedback gain as \(\bH \tilde{\bK} = \bK\), such that the resulting feedback law satisfies \(\bB \bK \delta\bx = \tilde{\bB} \tilde{\bK} \delta\bx\) and $\de \tbu = \tbK \de \bx$.

To correctly enforce the mutual exclusivity within the LQR synthesis via Riccati recursion, the state-feedback gain must be computed with respect to the minimal control input. Specifically, the control input matrix $\tbB$ is employed, and the gain $\tilde{\bK} \in \mathbb{R}^{2 \times 6}$, which operates on the independent variation \(\delta\tilde{\bu}\), is to be determined. This is structured as
\[
\tilde{\bK} = \begin{bmatrix} \tilde{\bK}_F^T & \tilde{\bK}_\delta^T \end{bmatrix}^T, \quad \tilde{\bK}_F, \tilde{\bK}_\delta \in \mathbb{R}^{1 \times 6}.
\]
The extended gain $\bK$ is automatically recovered via $\bK = \bH \tbK$ and, depending on the operating mode, it will result in the following expressions:
\begin{itemize}
  \item during \emph{acceleration}: \(\bK = \begin{bmatrix} \tilde{\bK}_F^T & \mathbf{0}^T & \tilde{\bK}_\delta^T \end{bmatrix}^T\),
  \item during \emph{braking}: \(\bK = \begin{bmatrix} \mathbf{0}^T & \tilde{\bK}_F^T & \tilde{\bK}_\delta^T \end{bmatrix}^T\).
\end{itemize}
This formulation preserves consistency with the complementarity constraint while retaining a structure amenable to controller synthesis.

A similar consideration applies when solving problem~\eqref{eq:ROCP} in Sec.~\ref{sec:CLrobustified}, where the complementarity structure must again be taken into account. In this case, it is sufficient to treat the elements of $\tbK_k$ as the minimal decision variables and reconstruct the full gain matrix $\bK_k$ via the mapping $\bK_k = \bH_k^{\text{S}} \tbK_k$, where
\begin{equation} \label{eq:minimalcontrolopt}
\bH^{\text{S}}(\tilde{\bu}) =
\begin{bmatrix}
\SIpos{u_1+u_2} & 0 \\
\SIneg{u_1+u_2} & 0 \\
0 & 1
\end{bmatrix}. 
\end{equation}
To ensure differentiability and numerical smoothness in the optimization problem~\eqref{eq:ROCP}, we employ smoothed approximations of the indicator functions $\Ipos{x}$ and $\Ineg{x}$, denoted as $\SIpos{x}$ and $\SIneg{x}$ in~\eqref{eq:minimalcontrolopt}, respectively. These are defined as:
\begin{equation} \label{eq:smoothedindicator}
\SIpos{x} :=
\frac{1}{2} \left(1 + \tanh(\chi x)\right), \quad \SIneg{x} := \frac{1}{2}\left(1 - \tanh(\chi x) \right),
\end{equation}
where the parameter $\chi>0$ controls the sharpness of the transition. Smaller values of $\chi$ yield smoother, more relaxed approximations, while larger values make the transition steeper and the function increasingly resemble a hard switching indicator.

%% file: performance_comparison.tex
\section{Performance comparison}
\label{sec:performance_comparison}

\begin{figure}
	\centering
	\includegraphics{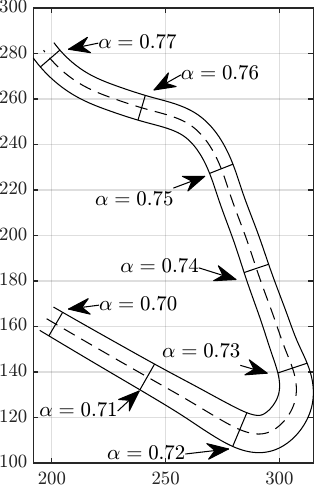}
	\caption{The analyzed Catalunya circuit sector spans $\alpha \in [0.70, 0.77]$, with labeled checkpoints every $\Delta\alpha = 0.01$. It contains two corners -- a low-speed turn ($\alpha = 0.72-0.73$) and a high-speed turn ($\alpha = 0.75-0.76$) -- providing contrasting dynamic conditions for vehicle behavior evaluation.}
	\label{fig:track}
\end{figure}

In this section we present some results obtained using the described robust planning frameworks. All planned trajectories refer to the sector of the Catalunya circuit shown in Fig.~\ref{fig:track}.
The entire circuit is parameterized by the curvilinear parameter $\alpha \in [0,1]$, and the selected sector corresponds to the interval $\left[0.70, 0.77\right]$. Checkpoints are indicated by labels and are uniformly spaced by $\De\al=0.01$. This sector includes two distinct corners: a low-speed turn from $\al = 0.72$ to $\al = 0.73$, and a high-speed turn from $\al = 0.75$ to $\al = 0.76$, allowing for the evaluation of vehicle behavior across different dynamic scenarios.

First we present selected results from the open-loop planning, which helps illustrate the method and its underlying mechanism.
This serves as a reference to assess the impact of key design parameters and constraint types on planned trajectories.
We then compare open- and closed-loop strategies, and finally validate the framework by comparing planned trajectories and simulations with noise realization.

The chosen sector was discretized into 140 spatial intervals ($\De s$ = 0.43 m). Both the open- and closed-loop disturbance-aware planning problems were transcribed on this grid in the CasADi-MATLAB environment~\cite{Andersson:MPC:2019}, and the resulting nonlinear programs were solved with IPOPT's interior-point algorithm~\cite{Wachter:MP:2006}.


\subsection{Open-loop parameter sensitivity}
\label{sec:ol_param_sensitivity}
To better understand the effect of key parameters, we focus on the open-loop formulation with robustified track limit constraint. This choice favors clearer visual interpretation.
The first parameter considered is the confidence level of constraint satisfaction.
We recall that the factor $\ga^\textrm{TLC}$ acts as a tuning knob, multiplying the standard deviation of the constraint, $\sigma^\textrm{TLC}$, to determine the total back-off term $\be^\textrm{TLC}$.
This parameter directly influences the probability of satisfying the track limit constraint: higher values of $\ga^\textrm{TLC}$ correspond to more conservative (i.e., robust) behavior.
Specifically, $\ga^\textrm{TLC} = 0$ yields a satisfaction probability of 50\% and leads to the nominal solution with no back-off, while $\ga^\textrm{TLC} = 3$ corresponds to a confidence level of 99\%.

\begin{figure}
	\centering
	\includegraphics{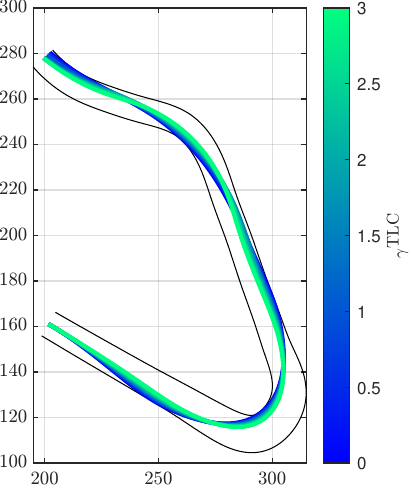}
	\hfill
	\includegraphics{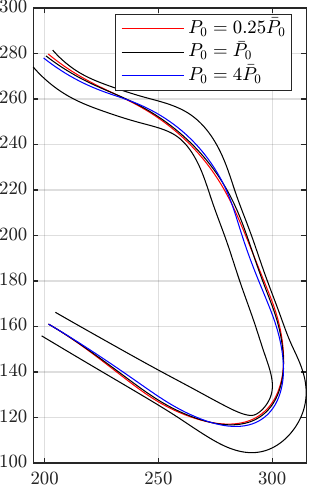}
\caption{Effect of the TLC penalty coefficient $\gamma^{\mathrm{TLC}}$ (left panel) and the initial covariance $\mathbf P_{0}$ (right panel) on the optimal trajectory.
Raising $\gamma^{\mathrm{TLC}}$ from 0 to 3 increases the probability of satisfying the track-limit constraints from 50 \% to 99 \%.
The right panel compares the nominal covariance $\bar{\mathbf P}_{0}$ (black) with $4\bar{\mathbf P}_{0}$ (blue line) and $\tfrac14\,\bar{\mathbf P}_{0}$ (red line), i.e., twice and half the nominal state uncertainty.
}
	\label{fig:ol_sensitivities}
\end{figure}

As shown in the left panel of Fig.~\ref{fig:ol_sensitivities}, trajectories with lower values of $\ga^\textrm{TLC}$ (darker blue) yield smaller lateral margins with respect to the track boundaries. In contrast, the trajectory with $\ga^\textrm{TLC} = 3$ (light green) remains noticeably farther from the edges.
This behavior is consistent with the increased conservativeness introduced by higher values of $\ga^\textrm{TLC}$, which amplify the back-off term $\be^\textrm{TLC}$ and thereby enforce a larger safety margin from the track limits.
Larger safety margins lead to higher sector times; spanning $\ga^\textrm{TLC}$ in the range $\left[0,3\right]$ results in sector times that vary from 12.135\,s to 13.162\,s.

The second parameter examined is the initial value of the covariance matrix $\bP_0$.
We recall that at each grid point a new prediction horizon is initialized, and the corresponding matrix $\bP$ at its starting point, i.e., $\bP^0_k$, is set to $\bP_0$.
The diagonal elements of $\bP$ are the variances $\Var{x_i}$  of each state variable, while the off-diagonal elements encode their covariances $\Cov{x_i}{x_j}$. Larger diagonal values indicate higher uncertainty in the corresponding states, and nonzero off-diagonal terms imply mutual dependence between them. For simplicity, a diagonal $\bP_0$ is used in this study, that is, for each prediction horizon we assign an initial standard deviation $\sig_{x_i}$ to each state and assume all initial covariances $\Cov{x_i}{x_j}$  to be zero.

 Assuming the state vector is ordered as described in Sec.~\ref{sec:vehicle_model}, the baseline initial covariance matrix is set to $\bP_0=\diag(\bbsig^2)$, with
\begin{align}
\bar{\boldsymbol{\sigma}}~=~\left[0.1\,\textrm{m/s}, 0.01\,\textrm{m/s}, 0.01\,\textrm{rad/s}, 1\,\textrm{m}, 1\,\textrm{m}, 0.0175\,\textrm{rad}\right]^T.
\end{align}
These standard deviations have been chosen based on the typical range of variation of each state. To provide a concrete interpretation at the starting condition, the assumption on first state $u \sim \calN(\mu_u,\sig_u^2)$, with $\sig_u=0.1$ m/s, implies that
$\Pr\{\mu_u-\sig_u \leq u \leq \mu_u+\sig_u\}\approx 68\%$ and $\Pr\{\mu_u-3\sig_u \leq u \leq \mu_u+3\sig_u\}\approx 99\%$.

In addition to the baseline value $\bar{\bP}_0$, two alternative values are considered: $\bP_0 = 4\bar{\bP}_0$ and $\bP_0 = \frac{1}{4}\bar{\bP}_0$. These correspond, respectively, to doubling and halving the initial standard deviations $\sig_{x_i}$ associated with each state vector.

A comparison between the trajectories obtained using different values of $\bP_0$ is shown in the right panel of Fig.~\ref{fig:ol_sensitivities}.
The black line corresponds to the baseline case with $\bP_0 = \bar{\bP}_0$, while the red and the blue lines represent the cases with $\bP_0 = \frac{1}{4}\bar{\bP}_0$ and $\bP_0 = 4\bar{\bP}_0$, respectively. As reasonably expected, using a larger initial covariance matrix results in an inflated covariance matrix that reflects increased uncertainty after $H$ steps, which in turn leads to a higher back-off term. This is associated to the blue line, which travels closer to the centerline with respect to the other two lines.

\subsection{Open-Loop vs. Closed-Loop comparison}

\begin{figure}[t]
	\centering
	\includegraphics{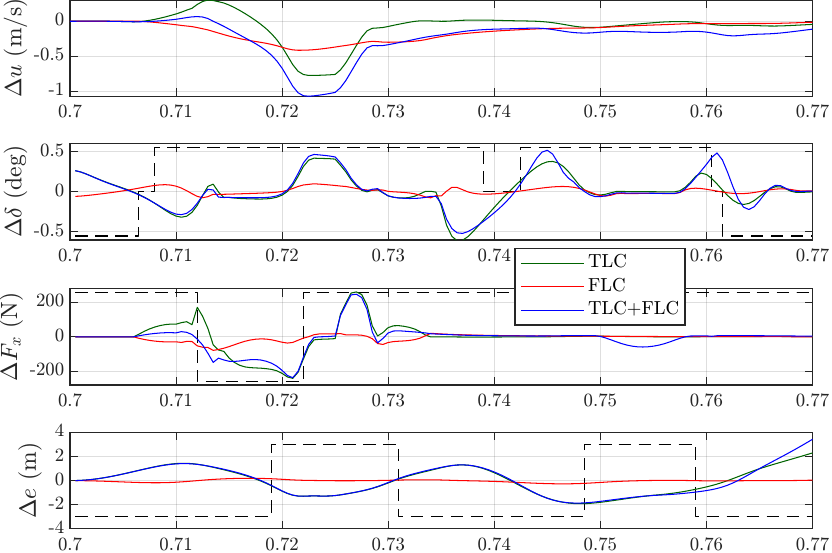}
	\caption{Open-Loop method. Variations of longitudinal speed $\Delta u$, steering angle $\Delta\delta$, longitudinal force $\Delta F_x$, and lateral deviation $\Delta e$ relative to the nominal trajectory for three schemes --- TLC (track-limit), FLC (friction-limit), and TLC+FLC (both).
	Dashed lines mark the sign of the reference signal (positive, near-zero and negative).}
	
	\label{fig:ol_telemetries}
\end{figure}

\begin{figure}[t]
	\centering
	\includegraphics[scale = 1]{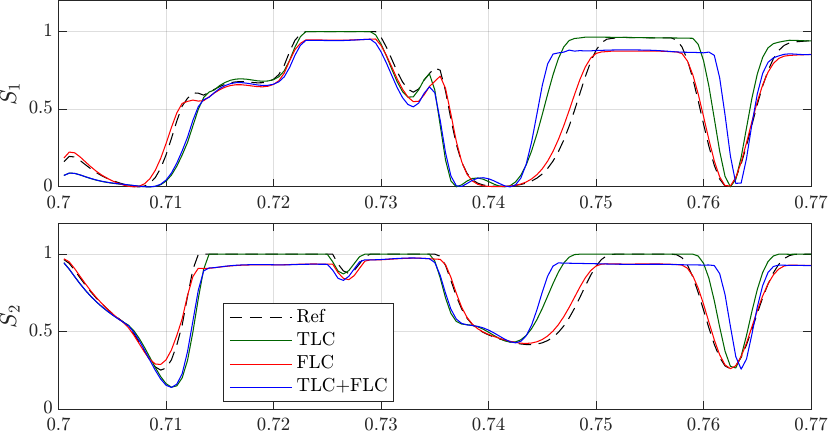}
	\caption{Open-Loop. Axle saturation $S_j(\bx,\bu)$ for front axle (j=1, first panel) and rear axle (j=2, second panel) with respect to the curvilinear parameter $\al$. Nominal configuration in dashed black, TLC in green, FLC in red, TLC+FLC in blue.
	}
	\label{fig:ol_saturation}
\end{figure}

We now investigate how the inclusion of different kinds of robustified constraints --- individually or in combination, as described in Sec.~\ref{sec:FLC} and~\ref{sec:TLC} --- influences the driving style.
We compare the nominal feed-forward optimal trajectory, obtained without robustified constraints, with those resulting from a robustified track limit constraint (denoted as TLC), a robustified friction limit constraint (denoted as FLC), and a scenario where both constraints are robustified (denoted as TLC+FLC).

First we show results obtained with the open-loop method described in Sec.~\ref{sec:open_loop_planning}, setting $H=4$, and $\ga^\textrm{TLC}=\ga^\textrm{FLC}=1.28$, corresponding to a 90\% probability of satisfying both constraints.
The comparison between the nominal case and the three robustified cases is illustrated in Fig.~\ref{fig:ol_telemetries}, in terms of the deviation of selected signals from to their nominal counterparts.
The panels display variations on the longitudinal speed $\De u$ (first panel), wheel steering angle $\De \de$ (second panel), total longitudinal force $\De F_x$ (third panel), and lateral deviation of the center of mass (CoM) from the track centerline $\De e$ (fourth panel).
Except for the $\De u$ panel, each plot includes dashed black lines that indicate the sign of the nominal signal.
These indicators are shown at three distinct vertical levels --- high, medium, and low --- corresponding respectively to positive, near-zero, and negative values of the nominal signal.


From the first panel, it can be observed that all robustified configurations, on average, show a lower longitudinal speed compared to the nominal trajectory, resulting in an increased sector time. Specifically, the configuration with the robustified TLC increases the sector time by 1.66\%, while that with the robustified FLC leads to a 0.79\% increase. When both TLC and FLC are robustified, the sector time increases by 2.57\%.
The two configurations incorporating the robustified TLC (blue and green lines) exhibit, as expected, a significantly altered CoM trajectory.
Indeed, the second and fourth panels clearly show that the TLC and TLC+FLC configurations tend to follow a path closer to the centerline. In particular, for most of the sector, the variation in lateral displacement $\De e$ and the lateral displacement $e$ itself exhibit opposite signs, indicating a corrective behavior of the steering angle $\de$ that pulls the vehicle toward the centerline.


This shift in trajectory and longitudinal speed becomes especially evident when approaching the sharp turn for $\al \in [0.707, 0.712]$. A noticeable increase in longitudinal speed is observed around $\alpha = 0.71$ for the TLC configuration  in the $\De u$ panel. This peak is followed by a sharp drop in speed, indicating intense braking.
The TLC+FLC configuration demonstrates a qualitatively similar trend in $\Delta u$, although the peak is significantly lower and the subsequent speed reduction is clearly stronger. This behavior is consistent with the variation of the total longitudinal force $\Delta F_x$ shown in the third panel.


Later in the sector, a further deviation in driving style appears during the high-speed turn, specifically around $\alpha \in [0.75, 0.76]$.
As shown in the third panel ($\Delta F_x$), in the combined configuration TLC+FLC a reduction in the longitudinal force within this sector is needed to meet both constraints.
This adaptation is necessary because the TLC forces the vehicle to follow a wider trajectory, which entails greater lateral acceleration and, consequently, higher lateral force demand. To satisfy the FLC under the resulting increase in tire utilization, the available accelerating force must be reduced.
Based on the nominal force sign indicator, this reduction is achieved by partially lifting the throttle.

After analyzing the variations of these four key signals with respect to the nominal solution, we now turn our attention to the evaluation of axle saturation $S_j(\bx,\bu)$ defined in Eq.~\eqref{eq:axle_saturation}, which provides insight into how close each configuration operates to the tire grip limits.

Figure~\ref{fig:ol_saturation} shows the values of $S_j(\bx,\bu)$ for the front axle ($j=1$, first panel) and the rear axle ($j=2$, second panel) for the previously introduced configurations: the nominal case (black dashed line) and the three robustified cases TLC, FLC and TLC+FLC (color-coded as before).

Firstly, we observe that the FLC and TLC+FLC configurations (red and blue lines) never reach full saturation ($S_j=1)$ due to the presence of the back-off term, which enforces a safety margin from the friction limit.
Secondly, we point out how the different trajectories followed by the TLC and TLC+FLC configurations (green and blue lines), compared to the nominal (black) and FLC case (red), entail distinct ground force demands, which are clearly visible here in terms of the $S_j$ ratio.
This behavior emerges around the sharp turn and becomes even more evident in the high-speed turn.
Specifically, in the interval $\al \in [0.745, 0.762]$, the TLC and TLC+FLC configurations begin to experience high axle loads approximately 12\,m before the nominal trajectory and sustain them for an additional 4.6\,m (TLC, green line) and 9.2\,m (TLC+FLC, blue line) beyond the nominal reference.


\begin{figure}[t]
	\centering
	\includegraphics{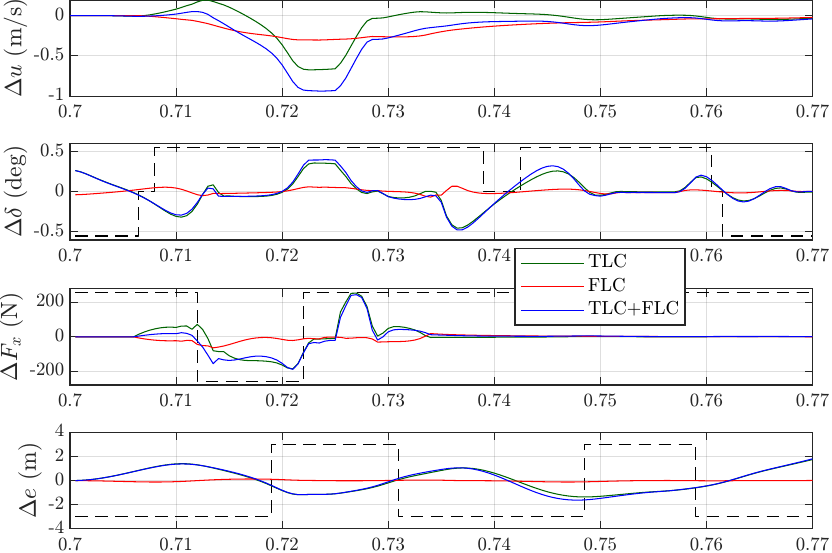}
	\caption{Closed-Loop method. Variations of longitudinal speed $\De u$, steering angle $\De\de$, longitudinal force $\De F_x$, and lateral deviation $\De e$ relative to the nominal trajectory for three schemes --- TLC (track-limit), FLC (friction-limit), and TLC+FLC (both).
		Dashed lines mark the sign of the reference signal (positive, near-zero and negative).}
	
	\label{fig:cl_telemetries}
\end{figure}
\begin{figure}[b!]
	\centering
	\includegraphics[scale = 1.005]{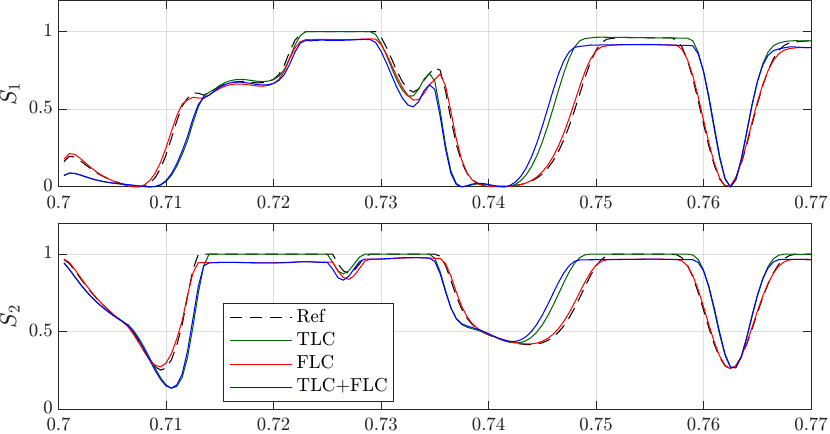}
	\caption{Closed-Loop. Axle saturation $S_j(\bx,\bu)$ for front axle (j=1, first panel) and rear axle (j=2, second panel) with respect to the curvilinear parameter $\al$. Nominal configuration in dashed black, TLC in green, FLC in red, TLC+FLC in blue.
	}
	\label{fig:cl_saturation}
\end{figure}

An analogous analysis has been conducted using the closed-loop approach described in Sec.~\ref{sec:closed_loop_planning}, setting $\ga^\textrm{TLC}=\ga^\textrm{FLC}=1.28$.
The same key signals displayed for the previous case are shown in Fig.~\ref{fig:cl_telemetries}, while the axle saturation ratios $S_j$ are illustrated in Fig.~\ref{fig:cl_saturation}.
From one side, the results appear very similar to those obtained with the open-loop method. While the closed-loop method offers a more rigorous framework by enabling seamless covariance propagation over the entire lap, the observed agreement indicates that open-loop planning remains a valid and effective strategy, as long as the prediction horizon is appropriately selected.

On the other hand, the trajectories are also sufficiently different to appreciate some subtle but relevant distinctions. For instance, sector time increments w.r.t. the nominal case of corresponding configurations are lower than the ones arising from the open-loop method: 1.39\% for the TLC, 0.66\% for the FLC, and 2.07\% for the TLC+FLC.
Moreover, in both TLC and TLC+FLC configurations, the magnitude of $\De\de$ is appreciably smaller in the closed-loop case compared to the open-loop one, especially during the high-speed turn.
Additionally, the TLC+FLC configuration no longer exhibits the throttle release in $\De F_x$ observed in the open-loop case.

As for the axle saturation, both methods lead to qualitatively similar evolutions, but slight differences can be observed within the high-speed turn interval $\al \in [0.745, 0.762]$. In this case, the TLC and TLC+FLC configurations begin to experience high axle loads approximately 10\,m before the nominal trajectory and sustain them for an additional 4.5\,m beyond the nominal reference. This is achieved through the closed-loop method's capability to \emph{steer} and \emph{tame} the covariance matrix propagation via the dynamic action of the controller.

\begin{figure}[b!]
	\centering
	\includegraphics{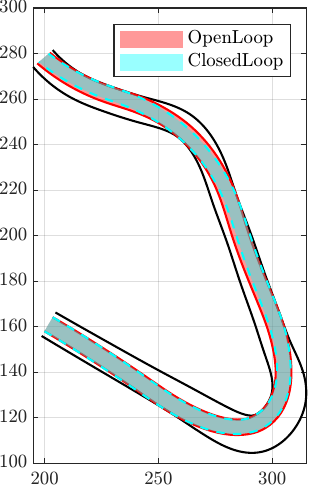}
	\caption{Back-off envelopes for the open-loop (red) and closed-loop (cyan) robust planners. Each shaded band represents the area swept by the back-off-extended vehicle and must lie entirely within the actual track boundaries.
		The open-loop approach yields a wider envelope due to worst-case growth assumptions, whereas the closed-loop design, aided by time-varying LQR controller, maintains a narrower margin.}
	\label{fig:olcl_traj_bands}
\end{figure}

Finally, we compare the influence of the back-off terms on the open-loop and closed-loop trajectories, in the robustified TLC scenario with the high $\ga^\textrm{TLC}=3.0$ value for better visualization. As before, the open-loop planning is carried out with a prediction horizon of $H=4$ steps.
The back-off term $\be^{TLC}$ in Eq.~\eqref{eq:TLC} admits an interesting geometrical interpretation: it corresponds to the amount by which the track boundaries are (not uniformly) tightened to meet the probabilistic constraint, or, equivalently, to a virtual extension of the vehicle's width.

According to the latter interpretation, Fig.~\ref{fig:olcl_traj_bands} shows two highlighted bands---red for the open-loop case and cyan for the closed-loop case---corresponding to the area swept by the back-off-extended-vehicle. Ultimately, the TLC enforces precisely that these bands remain entirely within the actual track limits. The comparison between the two regions illustrates how the additional conservativeness of the open-loop approach stems from larger back-off terms, i.e., a wider band that must stay within the track boundaries. The closed-loop method, on the other hand, benefits from the LQR-based feedback, which allows it to maintain smaller back-off terms and thus a narrower band, requiring a reduced safety margin from the track limits.

Solver times (build time is negligible) were measured on a laptop equipped with an Intel Core i9-13980HX with 32 GB RAM.
The reference (nominal) plan, described in Sec.~\ref{sec:nominalFF}, contains 4\,200 decision variables and is solved in 7.5\,s.
The open-loop robust plan (prediction horizon H = 4) described in Sec.~\ref{sec:open_loop_planning}, expands the model to 48\,300 variables, which IPOPT solves in 128\,s.
The closed-loop robust plan described in Sec.~\ref{sec:closed_loop_planning} first finds the nominal plan (4\,200 vars, 7.5\,s), executes a policy-search step (Step 2, Sec.~\ref{sec:LQR}, 0.5\,s), and then solves an additional robust problem with 14\,700 variables (Step 3, Sec.~\ref{sec:CLrobustified}, 33\,s), giving a total of 41\,s.

\subsection{Simulations with noise realization}
\begin{figure}[t!]
	\centering
	\includegraphics{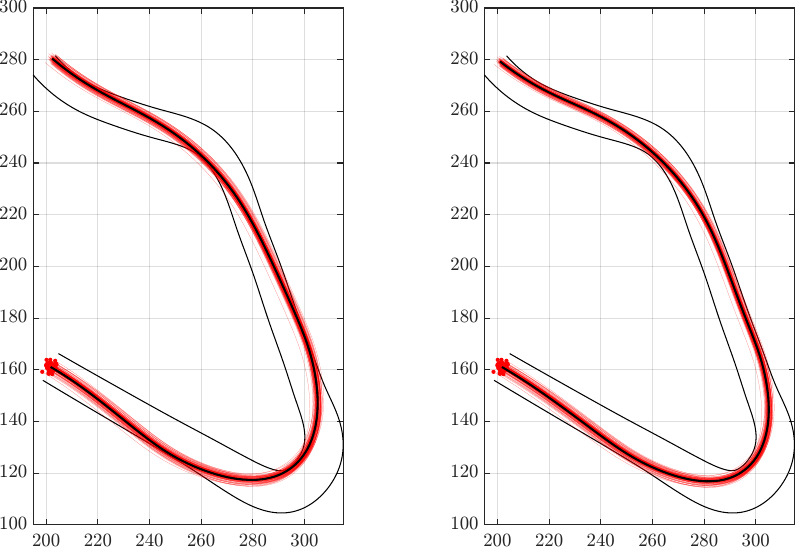}
	\caption{100 simulations (each panel) with random initial conditions and Gaussian noise realizations. In the left panel, the reference trajectory (black curve) is the non-robust one, tracked using an LQR controller designed around it as per Step 2 of Subsection~\ref{sec:LQR}. In the right panel, the reference (black curve) is a robust trajectory obtained using the closed-loop optimization method as per Step 3 of Subsection~\ref{sec:CLrobustified}, which is tracked using the corresponding optimized controller.}
	\label{fig:traj_strings}
\end{figure}
To provide an empirical validation of the closed-loop robustified planning we simulated the evolution of the controlled vehicle with 100 random initial conditions and included noise realizations. The random values for the initial state and the noise follow a Gaussian distribution with covariance matrices $\bP_0$ and $\bQ$, in both cases, as introduced in Eq.~\eqref{eq:dP}. This analysis is not applicable the open-loop approach, as it does not compute a stabilizing controller---resulting in trajectories that would significantly deviate from the track limits beyond the prediction horizon.

The left panel of Fig.~\ref{fig:traj_strings} shows random evolutions (red lines) for an LQR-controlled vehicle stabilized to follow the nominal optimal trajectory (black line). No robust constraints are introduced, and the controller implements the feedback policy from Step 2 of Subsection~\ref{sec:LQR}. The right panel, instead, shows the random evolutions (red lines) for the robustified LQR-based controller obtained in the Step 3 of the closed-loop approach described in Subsection~\ref{sec:CLrobustified}. The corresponding robustified mean trajectory is shown in black.
The comparison between the panels shows that the robustified controller is significantly more effective at satisfying the TLC, resulting in safer and more reliable trajectory tracking.



%% file: conclusions.tex
\section{Conclusions}
\label{sec:conclusions}

This work proposes and compares two robust planning strategies for minimum-time trajectory optimization in motorsport scenarios under uncertainty. The first approach, based on open-loop horizon-based covariance propagation, relies on a worst-case analysis of disturbance growth over a finite horizon to conservatively back off safety constraints. The second, a closed-loop covariance-aware method, incorporates a time-varying LQR feedback policy into the planning process, yielding a more realistic estimate of uncertainty evolution and tighter constraint enforcement.

Both schemes deliver reference trajectories suitable for human or artificial drivers. In autonomous applications the modelled controller can replicate the on-board implementation, providing highly accurate disturbance predictions; for human driving, fidelity improves with the extent to which the driver's behaviour can be approximated by the assumed time-varying LQR policy.

Quantitative results on a representative Barcelona-Catalunya sector highlight the trade-offs between robustness and performance. Open-loop planning, while conservative, ensures feasibility even in the absence of feedback, but incurs a higher performance cost due to larger safety margins. The closed-loop method achieves comparable safety guarantees with reduced conservativeness and improved lap-time performance, thanks to its ability to explicitly model the stabilizing action of the driver.

Comparing the computational costs of the two approaches, we observe that the closed-loop method demands significantly more programming effort but results in substantially lower total computation time. In contrast, the open-loop approach is simpler to implement but incurs higher solution times due to the larger number of optimization variables.

Finally, random simulations with noise realization, performed adopting the closed-loop approach, prove the effectiveness of the robustified constraints. A LQR-based controller tracking a robust trajectory planned with probabilistic constraint tightening exhibits significantly fewer track violations compared to one tracking the nominal optimal trajectory.

By embedding uncertainty propagation and feedback action directly in the planning stage, the proposed framework yields trajectories that are simultaneously fast and probabilistically safe, advancing minimum-time optimization toward practical deployment in high-performance motorsport and autonomous racing. Future work will explore the generalization to more accurate vehicle models and experimental validation of the methodology. 

%% file: MAIN.bbl
\begin{thebibliography}{10}
\providecommand{\url}[1]{\normalfont{#1}}
\providecommand{\urlprefix}{Available from: }

\bibitem{Veneri:FreetrajectoryQuasisteadystateOptimalcontrol:2020}
Veneri~M, Massaro~M. A free-trajectory quasi-steady-state optimal-control
  method for minimum lap-time of race vehicles. Vehicle System Dynamics. 2020
  Jun;\hspace{0pt}58(6):933--954.

\bibitem{Lovato:ThreedimensionalFixedtrajectoryApproaches:2022}
Lovato~S, Massaro~M. Three-dimensional fixed-trajectory approaches to the
  minimum-lap time of road vehicles. Vehicle System Dynamics. 2022
  Nov;\hspace{0pt}60(11):3650--3667.

\bibitem{Lovato:ThreedimensionalFreetrajectoryQuasisteadystate:2022}
Lovato~S, Massaro~M. A three-dimensional free-trajectory quasi-steady-state
  optimal-control method for minimum-lap-time of race vehicles. Vehicle System
  Dynamics. 2022 May;\hspace{0pt}60(5):1512--1530.

\bibitem{DalBianco:ComparisonDirectIndirect:2019}
Dal~Bianco~N, Bertolazzi~E, Biral~F, et~al. Comparison of direct and indirect
  methods for minimum lap time optimal control problems. Vehicle System
  Dynamics. 2019 May;\hspace{0pt}57(5):665--696.

\bibitem{Bertolazzi:DirectIndirectApproach:2025}
Bertolazzi~E, Biral~F. A {{Direct}}/{{Indirect Approach}} to~{{Optimal Control
  Problems}}. In: Sergeyev~YD, Kvasov~DE, Astorino~A, editors. Numerical
  {{Computations}}: {{Theory}} and {{Algorithms}}; Cham. Springer Nature
  Switzerland; 2025. p. 47--62.

\bibitem{Biniewicz:QuasisteadystateMinimumLap:2024}
Biniewicz~J, Pyrz~M. A quasi-steady-state minimum lap time simulation of race
  motorcycles using experimental data. Vehicle System Dynamics. 2024
  Feb;\hspace{0pt}62(2):372--394.

\bibitem{Bartali:SchwarzDecompositionParallel:2024}
Bartali~L, Gabiccini~M, Wright~SJ. Schwarz decomposition for parallel minimum
  lap-time problems: Evaluating against {{ADMM}}. Vehicle System Dynamics. 2024
  Sep;\hspace{0pt}:1--26.

\bibitem{Bartali:ConsensusbasedAlternatingDirection:2024}
Bartali~L, Grabovic~E, Gabiccini~M. A consensus-based alternating direction
  method of multipliers approach to parallelize large-scale minimum-lap-time
  problems. Multibody System Dynamics. 2024 Aug;\hspace{0pt}61(4):481--507.

\bibitem{Piccinini:HowOptimalMinimumtime:2024}
Piccinini~M, Taddei~S, Pagot~E, et~al. How optimal is the minimum-time
  manoeuvre of an artificial race driver? Vehicle System Dynamics. 2024
  Sep;\hspace{0pt}0(0):1--28.

\bibitem{Tapley:StatisticalOrbitDetermination:2004}
Tapley~BD, Schutz~BE, Born~GH. Statistical orbit determination. Amsterdam:
  Elsevier Academic Press; 2004.

\bibitem{Blackmore:ChanceConstrainedOptimalPath:2011}
Blackmore~L, Ono~M, Williams~BC. Chance-{{Constrained Optimal Path Planning
  With Obstacles}}. IEEE Transactions on Robotics. 2011
  Dec;\hspace{0pt}27(6):1080--1094.

\bibitem{Gao:CollisionfreeMotionPlanning:2023}
Gao~Y, Messerer~F, Frey~J, et~al. Collision-free {{Motion Planning}} for
  {{Mobile Robots}} by {{Zero-order Robust Optimization-based MPC}}. In: 2023
  {{European Control Conference}} ({{ECC}}); Jun.; Bucharest, Romania. IEEE;
  2023. p. 1--6.

\bibitem{Zhang:RobustifiedTimeoptimalPointtopoint:2025}
Zhang~S, Swevers~J. Robustified {{Time-optimal Point-to-point Motion Planning}}
  and {{Control}} under {{Uncertainty}}. arXiv. 2025
  Jan;\hspace{0pt}arXiv:2501.14526.

\bibitem{Zhang:RobustifiedTimeoptimalCollisionfree:2024}
Zhang~S, Bos~M, Vandewal~B, et~al. Robustified {{Time-optimal Collision-free
  Motion Planning}} for {{Autonomous Mobile Robots}} under {{Disturbance
  Conditions}}. In: 2024 {{IEEE International Conference}} on {{Robotics}} and
  {{Automation}} ({{ICRA}}); May; Yokohama, Japan. IEEE; 2024. p. 14258--14264.

\bibitem{Krog:SimpleFastRobust:2024}
Krog~H, J{\"a}schke~J. A simple and fast robust nonlinear model predictive
  control heuristic using n-steps-ahead uncertainty predictions for back-off
  calculations. Journal of Process Control. 2024 Sep;\hspace{0pt}141:103270.

\bibitem{Brault:RobustTrajectoryPlanning:2021}
Brault~P, Delamare~Q, Giordano~PR. Robust {{Trajectory Planning}} with
  {{Parametric Uncertainties}}. In: 2021 {{IEEE International Conference}} on
  {{Robotics}} and {{Automation}} ({{ICRA}}); May; Xi'an, China. IEEE; 2021. p.
  11095--11101.

\bibitem{Giordano:TrajectoryGenerationMinimum:2018}
Giordano~PR, Delamare~Q, Franchi~A. Trajectory {{Generation}} for {{Minimum
  Closed-Loop State Sensitivity}}. In: 2018 {{IEEE International Conference}}
  on {{Robotics}} and {{Automation}} ({{ICRA}}); May; Brisbane, QLD. IEEE;
  2018. p. 286--293.

\bibitem{Bohm:COPControlObservabilityaware:2022}
B{\"o}hm~C, Brault~P, Delamare~Q, et~al. {{COP}}: {{Control}} \&
  {{Observability-aware Planning}} ; 2022.

\bibitem{Meng:AnalysisGlobalCharacteristics:2022}
Meng~F, Shi~S, Zhang~B, et~al. Analysis for global characteristics of
  {{Lyapunov}} exponents in vehicle plane motion system. Scientific Reports.
  2022 Jun;\hspace{0pt}12(1):9300.

\bibitem{McCue:UseLyapunovExponents:2011}
McCue~LS, Troesch~AW. Use of {{Lyapunov Exponents}} to {{Predict Chaotic Vessel
  Motions}}. In: Almeida Santos~Neves~M, Belenky~VL, De~Kat~JO, et~al.,
  editors. Contemporary {{Ideas}} on {{Ship Stability}} and {{Capsizing}} in
  {{Waves}}. Vol.~97. Dordrecht: Springer Netherlands; 2011. p. 415--432.

\bibitem{Tamer:StabilityNonlinearTimeDependent:2016}
Tamer~A, Masarati~P. Stability of {{Nonlinear}}, {{Time-Dependent Rotorcraft
  Systems Using Lyapunov Characteristic Exponents}}. Journal of the American
  Helicopter Society. 2016 Apr;\hspace{0pt}61(2):1--12.

\bibitem{Cassoni:RotorcraftStabilityAnalysis:2024}
Cassoni~G, Cocco~A, Tamer~A, et~al. Rotorcraft stability analysis using
  {{Lyapunov}} characteristic exponents estimated from multibody dynamics. CEAS
  Aeronautical Journal. 2024 Jul;\hspace{0pt}15(3):703--719.

\bibitem{Sadri:StabilityAnalysisNonlinear:2013}
Sadri~S, Wu~C. Stability analysis of a nonlinear vehicle model in plane motion
  using the concept of {{Lyapunov}} exponents. Vehicle System Dynamics. 2013
  Jun;\hspace{0pt}51(6):906--924.

\bibitem{Gajic:LyapunovMatrixEquation:2010}
Gaji{\'c}~Z, Qureshi~MTJ, editors. Lyapunov matrix equation in system stability
  and control. San Diego: Academic Press; 2010. (Mathematics in Science and
  Engineering; v. 195).

\bibitem{Maruskin:DynamicalSystemsGeometric:2018}
Maruskin~JM. Dynamical {{Systems}} and {{Geometric Mechanics}}: {{An
  Introduction}}. Berlin/Boston: De Gruyter, Inc; 2018. (De {{Gruyter Studies}}
  in {{Mathematical Physics Ser}}; v.48).

\bibitem{Gillis:PracticalMethodsApproximate:2015}
Gillis~J. Practical methods for approximate robust periodic optimal control of
  nonlinear mechanical systems [dissertation]. Arenberg Doctoral School of
  Science, Engineering \& Technology, Dept. of Electrical Engineering, KU
  Leuven; 2015.

\bibitem{Andersson:MPC:2019}
Andersson~JAE, Gillis~J, Horn~G, et~al. {CasADi}: a software framework for
  nonlinear optimization and optimal control. Mathematical Programming
  Computation. 2019 Mar;\hspace{0pt}11(1):1--36.

\bibitem{Wachter:MP:2006}
W\"{a}chter~A, Biegler~LT. On the implementation of an interior-point filter
  line-search algorithm for large-scale nonlinear programming. Mathematical
  Programming. 2006 Mar;\hspace{0pt}106(1):25--57.

\end{thebibliography}
